\documentclass[11pt, twocolumn]{article}
\PassOptionsToPackage{hyphens}{url} 
\usepackage[margin=0.75in]{geometry}
\usepackage{authblk}
\usepackage{booktabs}
\usepackage{longtable}
\usepackage{array}
\usepackage{pdflscape}
\usepackage{xcolor}
\usepackage{amsmath}
\usepackage{hyperref}
\usepackage{microtype}
\usepackage{siunitx}
\usepackage{enumitem}
\usepackage{float}
\usepackage[normalem]{ulem}
\usepackage{graphicx}
\usepackage{subcaption}
\usepackage{xspace}
\usepackage{cleveref}

\definecolor{navy}{HTML}{173A5E}
\definecolor{good}{HTML}{1B7F4B}
\definecolor{warn}{HTML}{A45A00}
\hypersetup{
    colorlinks=true,
    linkcolor=navy,
    citecolor=navy,
    urlcolor=navy
}
\setlist{nosep,leftmargin=*}
\renewcommand{\arraystretch}{1.15}

\newcommand{\name}{HaCKRS\xspace}

\newcommand{\scaledmatrix}[1]{%
\begin{center}
\resizebox{0.98\linewidth}{!}{$#1$}%
\end{center}
}

\usepackage{booktabs}
\usepackage{tabularx}
\usepackage{wasysym}
\usepackage{makecell}

\usepackage{adjustbox}
\usepackage{caption} 

\newcommand{\featyes}{\CIRCLE}
\newcommand{\featpartial}{\LEFTcircle}
\newcommand{\featno}{\Circle}

\newcommand{\ie}{i.e.,\ }

\date{}

\title{A Paired RGB–Sonar Dataset for \\Fish Monitoring with Consumer Sonar}

\author[1, 2]{Mourad Derbedrosian*}
\author[1]{Michael A.~Hobley*}
\author[1]{Pietro Perona}

\affil[1]{California Institute of Technology \qquad \textsuperscript{2}UC Santa Barbara}
\affil[ ]{\small\texttt{mourad@ucsb.edu} \qquad \texttt{mahobley@caltech.edu}}

\begin{document}
\maketitle
{\renewcommand{\thefootnote}{\fnsymbol{footnote}}%
\footnotetext[1]{Joint first authors.}}

\begin{abstract}
Imaging sonar enables underwater observation in conditions where optical imaging is ineffective due to darkness and turbid water.
However, imaging-sonar systems previously used for ecological monitoring have typically been expensive, limiting their accessibility.
Consumer forward-looking sonar has the potential to make fish monitoring more accessible, but its noisy imagery makes annotation and the development of automated methods challenging.
We introduce the Huntington and Caltech Koi RGB and Sonar dataset (\name), which pairs Garmin LiveScope imaging-sonar recordings with top-down RGB video of koi carp.
The dataset contains more than 105,000 temporally matched frame pairs, spanning approximately 3.3 hours of recordings across four range and orientation configurations.
We provide frame-wise metadata, sonar geometry and coordinate information, synchronization data for all four configurations, and planar registration for the three horizontal configurations.
The release focuses on paired observations and calibration rather than task-specific fish annotations.
We intend \name to support annotation and the development of fish detection, tracking, and counting methods, and investigations into the potential of consumer sonar for ecological monitoring.
The \href{https://huggingface.co/datasets/perona-lab/Huntington_and_Caltech_Koi_RGB_and_Sonar}{dataset}, \href{https://github.com/fisheye-sonar/Huntington-and-Caltech-Koi-RGB-and-Sonar-Dataset}{code}, and \href{https://github.com/mahobley/OpenScope}{OpenScope}, the open-source tool we used to record the sonar, are publicly available.

\end{abstract}

\begin{figure}[!t]
 \centering
 \includegraphics[width=\linewidth]{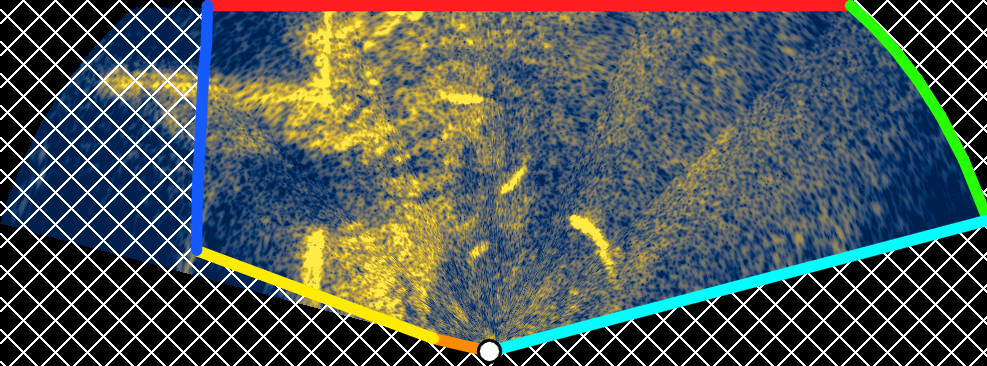}
 \includegraphics[width=\linewidth]{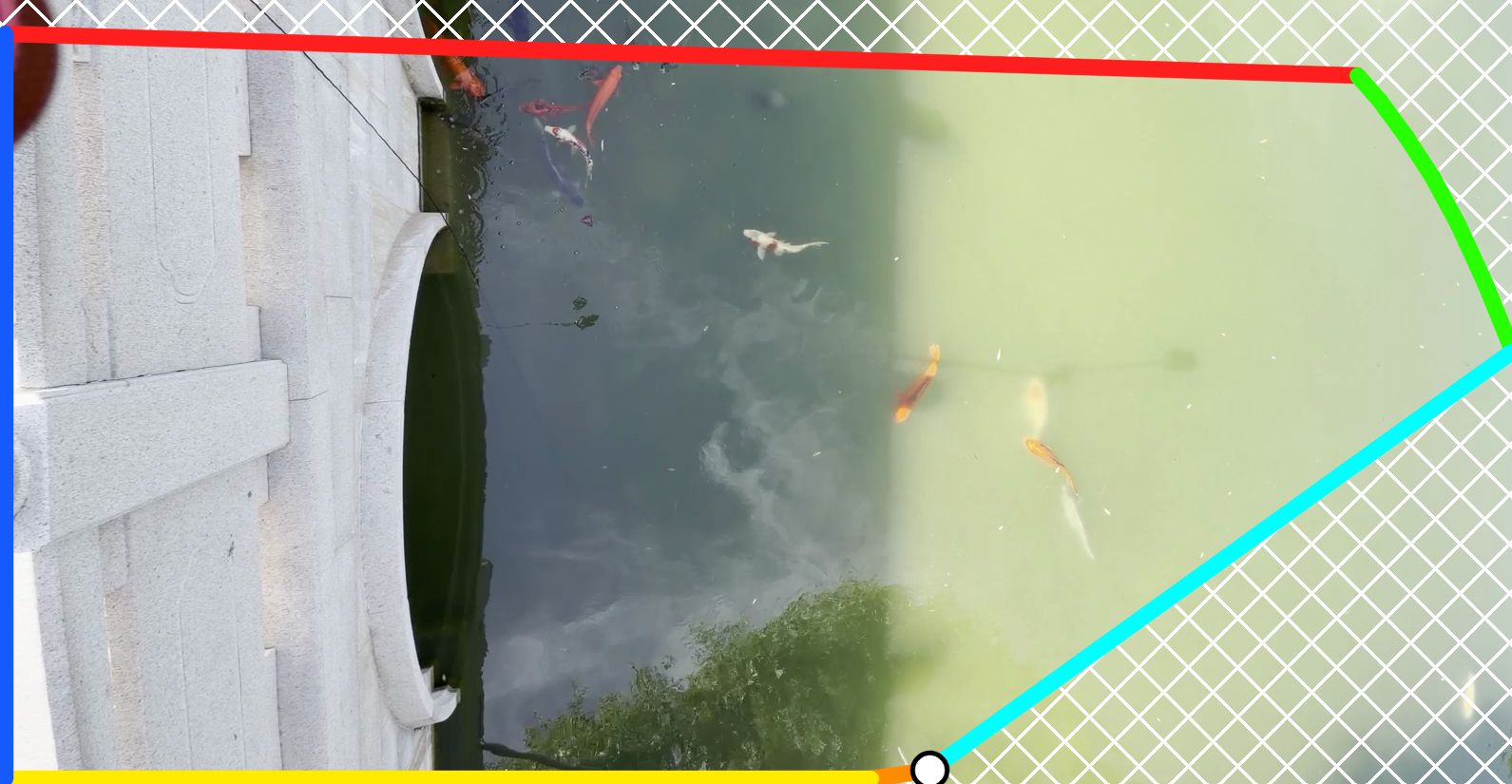}
 \caption{\name contains paired sonar and RGB frames of koi carp at the Huntington Gardens.
Colored lines correspond to matched boundaries between modalities.
The white circle denotes the sonar location.
Crosshatching denotes areas that were not captured by the other modality.}
 \label{fig:teaser}
\end{figure}
\section{Introduction}

Monitoring underwater environments is important for ecological science and fisheries management.
Video has clear utility for tasks such as fish detection, tracking, counting, and behavioral monitoring.

However, optical video is limited in many aquatic settings.
Its performance degrades in low-light or turbid conditions, restricting its use in natural settings.
Active sonar observes its environment by emitting acoustic signals and measuring their reflections off objects in the water, allowing it to operate in darkness, turbid water, poor weather, and at depth.
Imaging sonar measurements also contain metric range information and can, in principle, be used to estimate properties such as fish length and tail beat frequency.

These properties have motivated substantial work on fish detection, tracking, counting, and measurement using imaging sonar \cite{martignac2015didson,wei2022monitoring,munnelly2024applications}.
However, much of this work has focused on DIDSON and ARIS sonars, specialized and expensive scientific instruments.
These systems are often effective, but their cost significantly limits their accessibility and the scale at which they can be deployed.
Similar methods that could utilize lower-cost consumer sonar designed for recreational fishing could increase the scale at which environmental monitoring could be undertaken.

Developing computer-vision methods for these systems requires suitable training and evaluation data, yet sonar imagery can be difficult to interpret and annotate.
Fish returns may have poorly defined boundaries, overlap with environmental structure or other fish, and vary substantially with range, orientation, and scene geometry.

Synchronized optical imagery can provide complementary information when sonar returns are difficult to interpret.
A fish that is difficult to identify confidently in a sonar frame may be considerably easier to locate in a synchronized RGB observation.
Paired optical imagery can therefore assist annotation, model development, and diagnostic evaluation without requiring the optical sensor to remain available when a sonar-based method is ultimately deployed.

We recognize that planar transformations between RGB and sonar observations, as well as differences between the capture modalities, mean that perfect pixelwise correspondence cannot be expected.
However, the complementary information provided by the RGB imagery can still be valuable for guiding annotation and model development.

We introduce the Huntington and Caltech Koi RGB and Sonar (\textbf{\name}) dataset, consisting of paired imaging-sonar and top-down RGB recordings of koi carp at the Chinese Garden of The Huntington Library, Art Museum, and Botanical Gardens.
Sonar data was recorded from a Garmin Panoptix LiveScope LVS32 and GLS 10 system using OpenScope, our open-source sonar viewer and recorder, while the RGB imagery was captured using a smartphone.
The released paired imagery spans approximately 3.3 hours in four configurations: three horizontal sonar configurations at nominal ranges of 15, 40, and 100 feet, and one vertical configuration at 20 feet.
An example paired sonar and RGB observation is shown in \cref{fig:teaser}.

We temporally match the sonar and RGB frames for all four runs and estimate planar homographies between the two imaging modalities for the three horizontal configurations.
We also provide supporting information including frame-wise metadata, sonar coordinate transforms, and synchronization and calibration data; the dataset contents and statistics are summarized in \cref{sec:dataset_overview}.

\name is intended to support research on fish detection, tracking, and counting using accessible consumer imaging sonar.
Specifically, \name provides:
\begin{itemize}
   \item more than 105,000 temporally matched Garmin LiveScope sonar and top-down RGB frame pairs spanning approximately 3.3 hours
   \item four acquisition configurations spanning different sonar ranges and imaging-plane orientations
   \item frame-level timestamps, pairing information, and modality-specific acquisition metadata
   \item sonar geometry and coordinate information sufficient to reconstruct polar-to-Cartesian representations
   \item synchronization and calibration observations, including estimated planar RGB-sonar homographies for the three horizontal configurations.
\end{itemize}

The current release does not provide task-specific fish detections, identities, tracks, or count annotations.
Instead, it provides paired observations and geometric information from which such annotations and cross-modal methods can be developed.
In addition to providing the dataset itself, we document the practical synchronization, calibration, and acquisition issues encountered during collection; practical lessons from these are summarized in \cref{sec:lessons}.
We hope that these details make the resource easier to use and provide useful guidance for future multimodal sonar data collections.
The dataset, accompanying code, and OpenScope are publicly available; links are given in \cref{sec:data_availability}.

\section{Related work}
\subsection{Imaging sonar for fish monitoring}

Imaging sonar has been widely used for observing fish in environments where optical cameras are limited by turbidity, low light, or occlusion. Systems such as DIDSON and ARIS provide high-resolution acoustic imagery and have been applied to fish detection, counting, tracking, and behavioral monitoring \cite{martignac2015didson,wei2022monitoring,munnelly2024applications}. These systems have also motivated work on automated analysis of sonar imagery, including methods for detecting and counting individual fish \cite{connolly2023shadows,duan2023boosting,shen2024didson}. The Caltech Fish Counting Dataset (CFC22) provides a large-scale benchmark for fish detection, tracking, and counting in imaging-sonar video \cite{cfc2022eccv}, while CFC26 extends this work toward evaluations relevant to deployment \cite{cfc2026neurips}. SAVeD further demonstrates how self-supervised denoising of low-SNR video can improve downstream detection, tracking, and counting performance on sonar imagery \cite{stathatos2026saved}.
However, scientific imaging-sonar systems can require specialized hardware and deployment procedures, motivating interest in more accessible sonar platforms.

More recently, consumer forward-looking sonar systems have begun to be explored for ecological monitoring. Dickson et al.~\cite{dickson2026sonar} adapted a Garmin Panoptix LiveScope system for ecological monitoring in the Dutch Wadden Sea. They mounted the system on a customized floating platform and used it to measure pelagic fish abundance around experimental tree-reef and paired control sites. Their study demonstrates the potential of consumer fish-finding sonar for non-invasive ecological monitoring in shallow, turbid environments where conventional optical surveys can be difficult.

This growing use of consumer imaging sonar motivates the development of datasets and analysis tools tailored to these systems. Whereas prior ecological work has demonstrated that LiveScope can provide useful measurements of fish abundance \cite{dickson2026sonar}, our work focuses on the data and methodological infrastructure needed to support automated interpretation of its imagery. Specifically, \name provides temporally paired LiveScope and RGB observations, together with sonar geometry, synchronization information, and approximate spatial registration, enabling the study of fish observations across the two sensing modalities.

\noindent\textbf{Joint optical and acoustic observation of fish.}
Optical imagery has been combined with imaging sonar to validate acoustic observations and provide complementary biological information. In the Ocqueoc River dataset, McCann et al.\ used synchronized underwater video to identify fish observed by DIDSON and produce high-confidence labeled clips \cite{mccann2018underwater}. Simultaneous camera and sonar recordings have also been used for cross-modal image generation in aquaculture and for direct comparisons of fish detectability and density estimates between acoustic and optical systems \cite{terayama2019integration,sibley2023sound}. More recently, the HODOR system combines forward-looking imaging sonar with low-light stereo cameras for continuous ecological observation and provides hundreds of hours of synchronized sonar and optical recordings \cite{groger2024observatory,wilts2025hodor}. Collectively, these studies demonstrate the value of complementary acoustic and optical observations. They also motivate paired datasets in which the more readily interpretable optical modality can assist the analysis of sonar imagery.

\noindent\textbf{Paired sonar--visual datasets and accessible sensing.}
Paired sonar and visual data has also been developed for broader underwater perception tasks. RGBS50 contains temporally aligned RGB-sonar sequences for single-object tracking \cite{li2025rgbsonar}. Dahn et al.\ provide more than 74,000 matched ARIS and GoPro observations of underwater unexploded ordnance together with calibration and pose information \cite{dahn2024acoustic}. Other AUV and ROV datasets combine forward-looking sonar with camera imagery and navigation measurements. SOLAQUA provides camera, sonar, and navigation data collected at an operational fish farm \cite{ohrem2025solaqua}, while SOVIS provides more than 76,000 paired sonar--visual frames from open-water ROV deployments and demonstrates cross-modal fish detection \cite{chen2026sovis}. These datasets establish temporal pairing and multimodal acoustic--visual sensing in scientific and robotic settings.

In parallel, consumer forward-facing sonar systems such as Garmin LiveScope have begun to be evaluated for ecological monitoring, suggesting that sensors originally developed for recreational fishing may also be useful for scientific observation \cite{dickson2026sonar}. \name occupies a complementary setting: it pairs observations of fish from a consumer LiveScope system with a top-down consumer RGB camera. All four acquisition configurations provide temporally matched RGB-sonar observations and explicit sonar geometry, while the three horizontal configurations additionally provide planar RGB-sonar mappings. This combination is intended to support the development and evaluation of automated fish-monitoring methods specifically for accessible consumer imaging sonar.

A structured comparison of \name with these related datasets and studies is provided in Table~\ref{tab:related_datasets} in the Appendix.

\section{Dataset Overview}
\label{sec:dataset_overview}

\name contains paired sonar and top-down RGB observations of koi carp recorded at a single pond in the Chinese Garden of The Huntington Library, Art Museum, and Botanical Gardens.
The sonar and camera remained in fixed acquisition configurations during each run while fish moved freely through the observed region.

\begin{figure}[!t]
 \centering

 \begin{subfigure}{\linewidth}
     \centering
     \includegraphics[width=\linewidth]{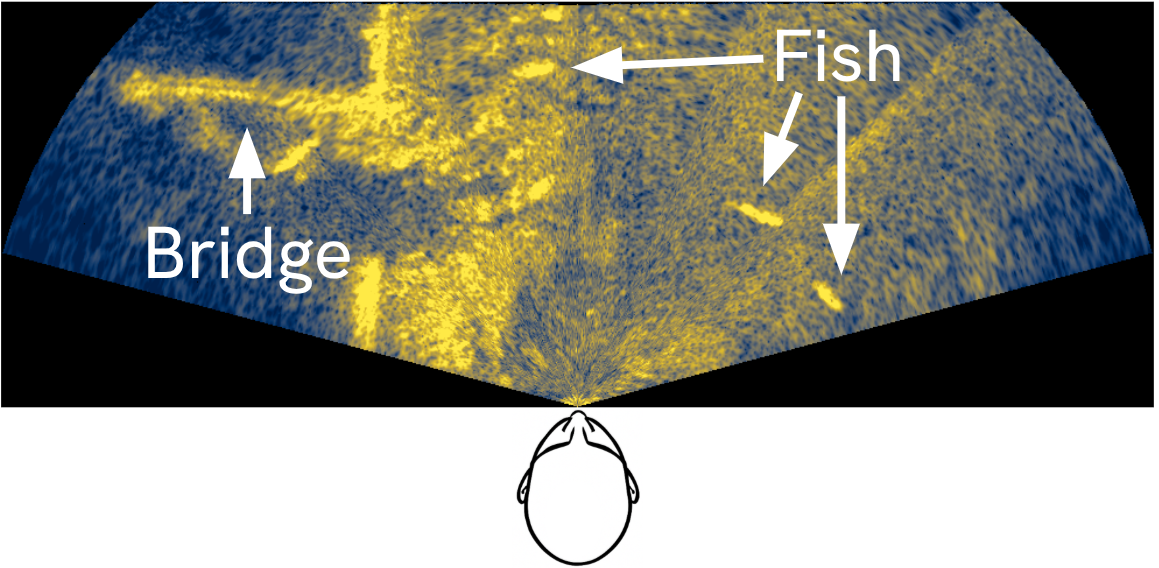}
     \caption{Horizontal orientation (15H, 40H, 100H).}
     \label{fig:viewpoint-horizontal}
 \end{subfigure}

 \vspace{1mm}

 \begin{subfigure}{\linewidth}
     \centering
     \includegraphics[width=\linewidth]{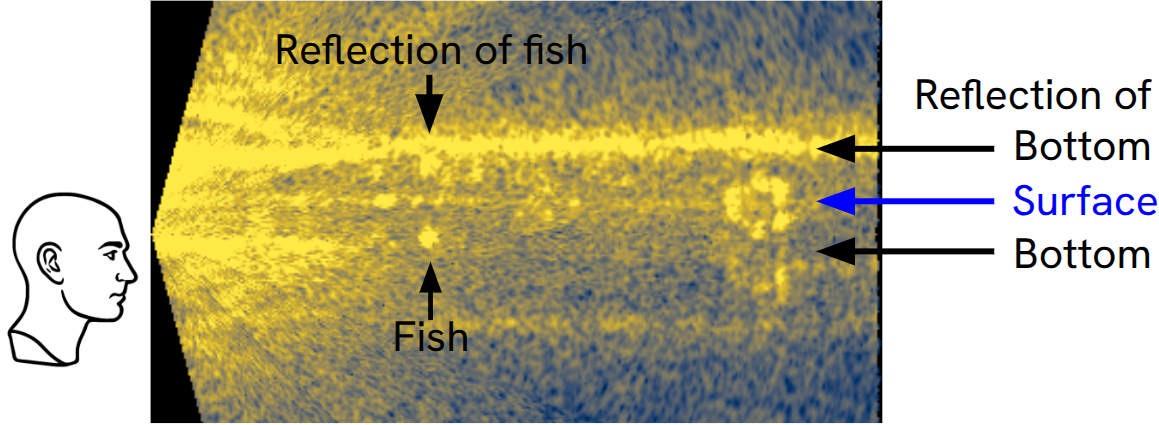}
     \caption{Vertical orientation (20V) cropped.}
     \label{fig:viewpoint-vertical}
 \end{subfigure}

 \caption{\looseness=-1 Horizontal and vertical viewpoints.
The horizontal sonar approximates a top down view.
The vertical run takes a thin profile slice through the water column.
In the vertical setting, the sensor is near the pond floor. Significant reflections can be seen from both the water surface and the hard pond floor.}
 \label{fig:viewpoint}
\end{figure}

The dataset contains four configurations, denoted 15H, 40H, 100H, and 20V.
In 15H, 40H, and 100H, the sonar imaging plane was oriented approximately horizontally, producing a plan-like view that can be related to the top-down RGB camera.
These configurations used nominal sonar ranges of 15 feet ($\sim$4.5m), 40 feet ($\sim$12m), and 100 feet ($\sim$30m), respectively.
In 20V, the sonar was rotated 90 degrees about its centerline and placed near the pond floor, producing a vertical profile through the water column at a nominal range of 20 feet ($\sim$6m).
The horizontal and vertical sonar orientations are illustrated in \cref{fig:viewpoint}.

The selected release intervals exclude frames containing people, the calibration object, and the manual synchronization actions.

\subsection{Dataset contents}

The release is intended to preserve both the observations themselves and the information required to interpret their temporal and spatial relationship.
RGB frames are decoded from MOV files, while sonar frames are extracted as timestamped PNGs whose identifiers may contain gaps.

For each paired observation, we provide the corresponding sonar and RGB frame identifiers and timestamps together with their temporal difference.
Sonar metadata includes the recorded frame identifier, wall-clock timestamp, source clip, polar image dimensions, range information, and supplied angular coordinates.
RGB metadata includes the corresponding frame identifier, timestamps, source capture, image dimensions, and available camera metadata.
The frame-wise metadata fields are detailed in \cref{sec:framewise_metadata}.

For the sonar modality, we retain the geometry needed to interpret the recorded polar representation and construct Cartesian representations.
The Cartesian transformation used in this work is described in Sec.~\ref{sec:sonar_cartesian}, but the release retains the underlying coordinate information so that users are not restricted to our particular rasterization or interpolation procedure.

The release also includes the observations used to establish and assess correspondence between the modalities.
Manual synchronization events, in which a physical object was rapidly introduced into the scene and observed in both modalities, are retained separately from the matched analysis footage.
Frames containing the geometric calibration object are similarly retained as supporting calibration data.

For the three horizontal configurations, we provide the estimated planar homography between the Cartesian sonar representation and the corresponding RGB view together with the calibration correspondences used to estimate it.
The corresponding temporal-alignment and spatial-calibration procedures are described in \cref{sec:temporal_alignment,sec:homography}.
No planar homography is reported for the vertical 20V configuration because its geometry and available correspondences did not support a defensible planar mapping.

The release therefore distinguishes between three types of information: the primary paired observations, acquisition and pairing metadata describing those observations, and supporting synchronization and calibration data documenting how the correspondence was obtained.

\subsection{Dataset statistics}

\Cref{tab:run_summary} summarizes the selected recording intervals and matched frame counts for the four configurations.
Together, these intervals span approximately 3.3 hours and contain 105,362 matched RGB-sonar pair records.
Each pair contains a unique sonar frame.
Four RGB frames are each associated with two sonar frames near recording boundaries, resulting in 105,358 unique RGB frames in the paired release; these cases are listed in \cref{tab:duplicate-boundary-matches}.
Matched RGB-sonar image pairs have less than 100~ms temporal discrepancy, with the median difference below 8.5~ms. Per-run temporal differences are summarized in \cref{tab:temporal_difference_for_matches}.
Sonar frame-rate statistics are summarized in \cref{tab:sonar_fps}, with frame-rate variation over time and its distribution shown in \cref{fig:fps,fig:violin}.

\begin{table*}
\centering
\newcommand{\range}[2]{%
 \makebox[2.5em][r]{#1ft}%
 \hspace{0.1em}/\hspace{0.1em}%
 \makebox[3.0em][r]{#2m}%
}
\begin{tabular}{rccccc}
\toprule
Run & Range & Orientation & Duration & Total Frames & Matched Pairs\\
\midrule
15H  & \range{15}{4.57}   & Horizontal & 1:03:08.018 & 36,028 & 35,984 \\
40H  & \range{40}{12.19}  & Horizontal & 1:09:11.831 & 32,898 & 32,878 \\
100H & \range{100}{30.48} & Horizontal & 0:47:25.255 & 20,641 & 20,635 \\
20V  & \range{20}{6.10}   & Vertical   & 0:18:18.507 & 15,869 & 15,865 \\
\bottomrule
\end{tabular}
\caption{Recording durations and matched frame counts by run.
Sonar frames were matched with the nearest RGB frame. Frames without an RGB observation within 100~ms were excluded from the paired release.
Duration is in HH:mm:ss.fff.}
\label{tab:run_summary}
\end{table*}

\begin{table}
\centering
\small
\begin{tabular}{rccc}
\toprule
Run & RGB frame \# & Sonar frame \# & $\Delta t$ (ms)\\
\midrule
15H & 50,960 & 50,657 & $+82.63$\\
15H & 50,960 & 50,658 & $+27.27$\\
\midrule
40H & 231,208 & 25,589 & $+9.08$\\
40H & 231,208 & 25,590 & $-67.29$\\
40H & 231,230 & 25,599 & $+73.08$\\
40H & 231,230 & 25,600 & $+3.35$\\
\midrule
20V & 438,245 & 133,025 & $-13.23$\\
20V & 438,245 & 133,026 & $-76.86$\\
\midrule
100H & \multicolumn{2}{c}{All matches are one-to-one} & ---\\
\bottomrule
\end{tabular}
\caption{Cases where the same RGB frame was matched to two sonar frames.
These occur around RGB capture boundaries where fewer RGB frames were captured due to file saving.
Overall, \name contains 105,362 pair records for the 105,362 unique sonar frames linked to 105,358 unique RGB frames.
A positive difference means the corrected RGB timestamp is later than the sonar timestamp.}
\label{tab:duplicate-boundary-matches}
\end{table}

\begin{table}
\centering
\begin{tabular}{rcccc}
\toprule
Run & Median & 95th\% & Max & $\leq$33 ms \\
\hline
15H & 8.30 ms & 15.86 ms & 82.63 ms & 99.989\% \\
40H & 8.67 ms & 15.86 ms & 91.91 ms & 99.982\% \\
100H & 8.20 ms & 15.88 ms & 63.12 ms & 99.990\% \\
20V & 8.05 ms & 15.85 ms & 76.86 ms & 99.987\% \\
\bottomrule
\end{tabular}
\caption{Temporal difference between the matched RGB and sonar frames for each run.
Frames with $>$100~ms difference were not matched.
95th\% refers to the 95th percentile.}
\label{tab:temporal_difference_for_matches}
\end{table}

\begin{table}
\centering
\begin{tabular}{rrrrr}
\toprule
& \multicolumn{4}{c}{Frames per Second}\\
\cmidrule(lr){2-5}
Run & 5th\% & Mean &  Median &  95th\% \\
\hline
15H & 6.2 & 9.50 & 8.6 & 15.2 \\
40H & 5.0 & 7.92 & 6.4 & 13.4 \\
100H & 5.4 & 7.25 & 7.6 & 8.6 \\
20V & 14.0 & 14.44 & 14.4 & 15.0 \\
\bottomrule
\end{tabular}
\caption{Frames per second of the sonar frames.
5th\% and 95th\% refer to the 5th and 95th percentiles.
See \cref{fig:fps} for a graph of how the frame rate changes over the duration of each run, and see \cref{fig:violin} for a visualization of the frame rate distribution.}
\label{tab:sonar_fps}
\end{table}

\subsection{Intended use}

\name is intended primarily as a data resource for studying automated analysis of recreational imaging sonar.
Potential uses include fish detection, tracking, and counting; development of annotation procedures for difficult sonar imagery; analysis of range-dependent sonar appearance; and cross-modal methods in which optical observations provide additional supervision during development.

The paired RGB imagery should not be interpreted as exhaustive ground truth for the sonar modality.
Instead, it provides an additional observation of the same scene with different geometry and different failure modes.
Methods using \name should account explicitly for incomplete field-of-view overlap, depth-dependent geometric disagreement, temporal uncertainty, and cases in which an object is visible in only one modality.
These limitations are discussed in more detail in \cref{sec:limitations}.

\section{Data Collection and Alignment}
We recorded sonar and top-down RGB imagery for a total of 3.3 hours.
Recordings were conducted at a single location in the Chinese Garden at The Huntington Library, Art Museum, and Botanical Gardens.

\subsection{Recording specifications}

RGB imagery was captured as video using the ``Ultra Wide'' lens on an iPhone 17 at a resolution of 1920$\times$1080 pixels and 30 frames per second.

Sonar imagery was recorded using a Panoptix LiveScope LVS32 transducer and GLS 10 sonar module.
Sonar frames and their metadata were recorded with OpenScope, our open-source desktop viewer and recorder for the GLS 10 and LVS32.
The imagery is composed of 511 angular beams and  512 range bins.
The extracted metadata contains 511 angular coordinates spanning approximately 150°. This differs from Garmin's specified 135° front-to-back field of view~\cite{garmin2021livescope}. We report the recorded coordinate values here without assuming that their full numerical span corresponds directly to the transducer's effective physical field of view.
The recorded angular spacing is characterized in \cref{appendix_angular_spread}.

The recorded polar imagery is incomplete at the largest ranges: the first 345--351 range bins span the full angular range (see \cref{tab:num_complete_polar_rows} for the variation across runs). Beyond these bins, the angular coverage is incomplete. When transformed into Cartesian coordinates, this produces the flat upper edge shown in \cref{fig:polarcartesian}.
The straight horizontal edge has a constant \(y\)-coordinate in the Cartesian representation at the specified range.
When we refer to the range of a video, we mean the range at the centerline.
The advertised beam height for the sensor is 20 degrees~\cite{garmin2021livescope}, though we were not able to confirm this. The returned values appear to be a combination of all the reflections in the 20 degree beam. This can increase the apparent size of objects as they can intersect multiple range bins in the same beam.

Both modalities were saved in clips at 15 minute intervals.
Clip-level sonar and RGB inventories are provided in \cref{tab:sonar_inventory,tab:rgb_inventory}.
After recording, they were combined into a complete corpus of frames.

\subsection{Sonar orientation}
Three of the runs (15H, 40H, and 100H) were conducted with the sonar imaging plane oriented horizontally, approximating a top-down view.
For these runs, the sensor was positioned less than 15~cm (approximately 6~inches) below the water surface.
In the vertical run (20V), the sonar was rotated 90 degrees about its centerline, giving a vertical or profile view, and placed in contact with the pond floor, approximately 60~cm (2~feet) below the surface. 

\subsection{Temporal alignment}
\label{sec:temporal_alignment}
We temporally aligned the frames using each capture mode's internal frame-wise metadata.
To check our temporal alignment, we performed an action that was visible in both modalities. 
We quickly placed a pipe vertically into the water. 
This produced a suddenly appearing bright area in the sonar imagery.

RGB timestamps were converted to a common wall-clock timeline and aligned with the sonar timestamps. At some file-save boundaries, the RGB stream briefly froze or repeated frames, creating small discontinuities in the timeline. Separate timing corrections were applied on either side of these discontinuities to preserve alignment with the continuously recorded sonar stream.
Additional details of the timestamp correction and synchronization quality are provided in \cref{sec:appendix_time_sync}.

Frames that included the manual time-sync actions are not included in the released temporally matched dataset, but are included as additional data.

The sonar frame rate sometimes decreased over the course of a recording, for reasons we have not determined. These variations are documented in \cref{sec:appendix_frame_rate_decrease}. Temporal matching therefore uses the recorded frame timestamps rather than assuming a constant sonar frame rate.

Frames were paired by finding the temporally closest RGB frame to each sonar frame. 
If no RGB frame was found within 100~ms, no match was made and the sonar frame was excluded from the dataset.

Of the 105,436 selected sonar frames before temporal matching, 105,362 were paired with an RGB frame within 100~ms.
The remaining 74 were excluded. 
All 105,362 matches contain a unique sonar image. Four RGB frames are each paired with two sonar frames near recording boundaries, so the released dataset includes 105,358 unique RGB frames.
The duplicate boundary matches and per-run temporal differences are reported in \cref{tab:duplicate-boundary-matches,tab:temporal_difference_for_matches}.

\subsection{Sonar polar-to-Cartesian transformation}\label{sec:sonar_cartesian}
\begin{figure*}
 \centering
 \subcaptionbox{Raw polar sonar frame\label{fig:polarcartesian_raw}}{
     \includegraphics[height=44mm]{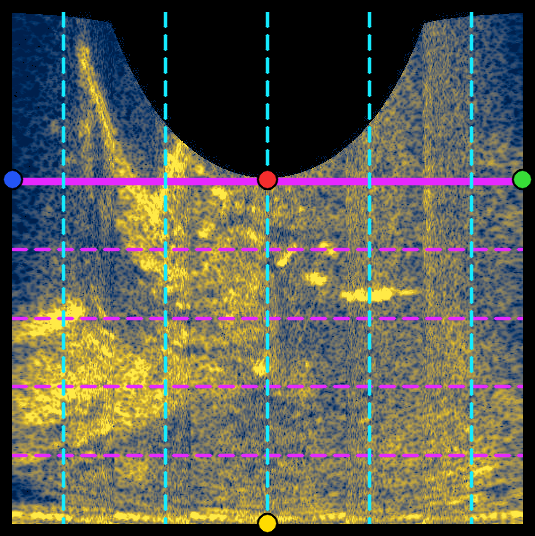}
 }
 \hfill
 \subcaptionbox{Cartesian-mapped sonar frame\label{fig:polarcartesian_warped}}{
     \includegraphics[height=44mm]{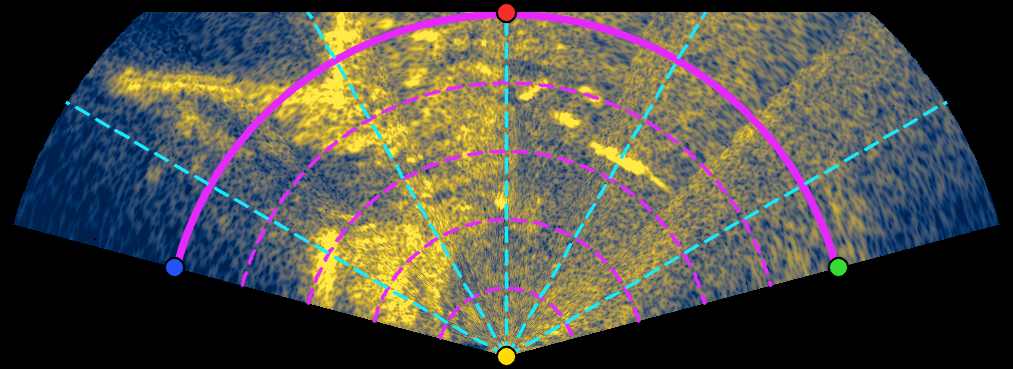}
 }
 \caption{Example imaging-sonar frame before and after cartesian mapping.
 The same 3 ft (91 cm) range lines and 30$^\circ$ angular reference lines are shown in both.}
 \label{fig:polarcartesian}
\end{figure*}

To support geometric comparison between the two modalities, we transform the recorded sonar measurements from their polar representation into a Cartesian coordinate system, see \cref{fig:polarcartesian}.
The Cartesian representation also provides a uniform spatial coordinate system in the sonar imaging plane, making geometric comparisons more convenient, \ie objects of the same size should appear the same size.
For this transformation, we used the beam angles and bin distances defined in the recorded sonar metadata.

For the visualization in this paper we used inverse mapping to convert each Cartesian output pixel into its corresponding range and beam-angle coordinates in the polar image. 
We then used bilinear interpolation between the neighboring polar samples to determine its intensity. 
Pixels outside the valid recorded range and angular extent were assigned a value of zero.
The dataset includes the original polar imagery and the underlying sonar coordinate information so that alternative representations can be constructed.

\subsection{RGB-sonar planar registration}\label{sec:homography}
\begin{figure*}
 \centering

 \begin{subfigure}[t]{\textwidth}
     \centering
     \includegraphics[height=26mm]{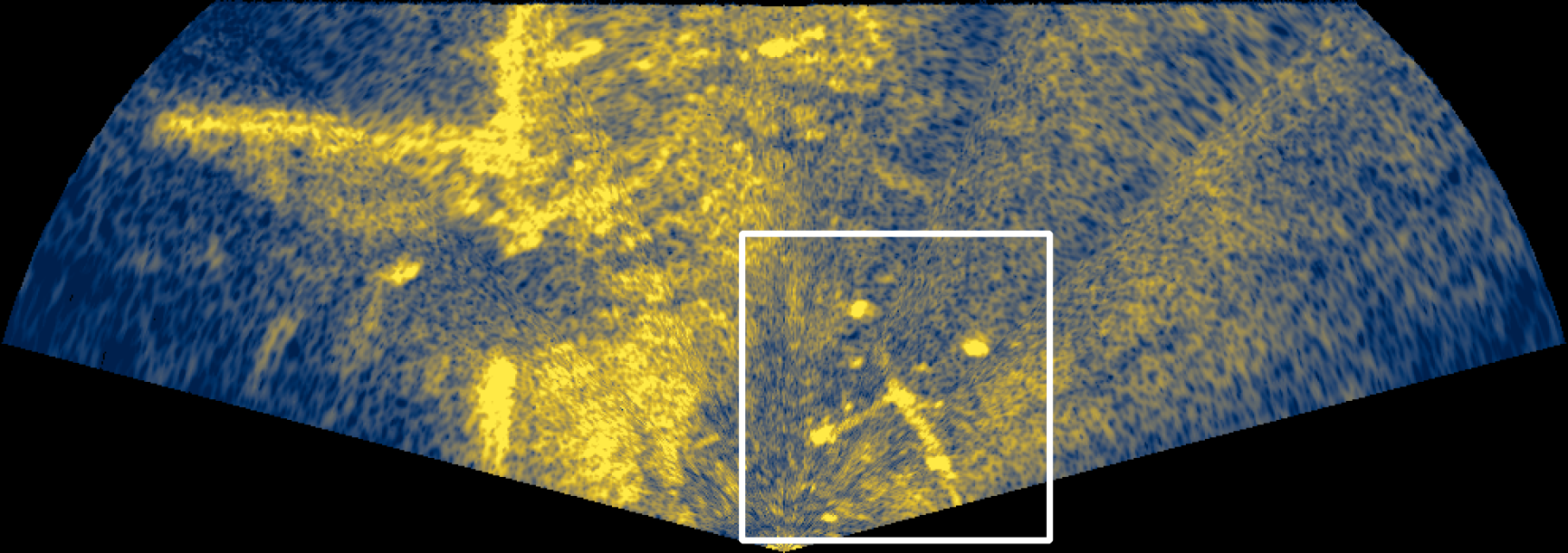}
     \hfill
     \includegraphics[height=26mm]{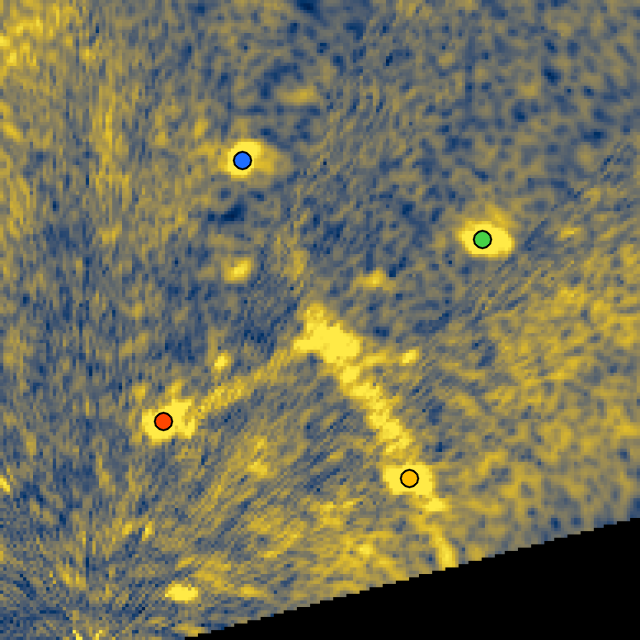}
     \hfill
     \includegraphics[height=26mm]{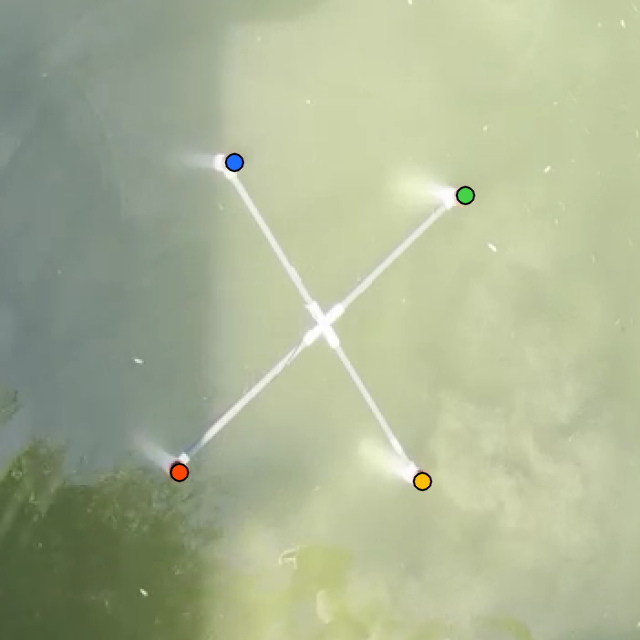}
     \hfill
     \includegraphics[height=26mm]{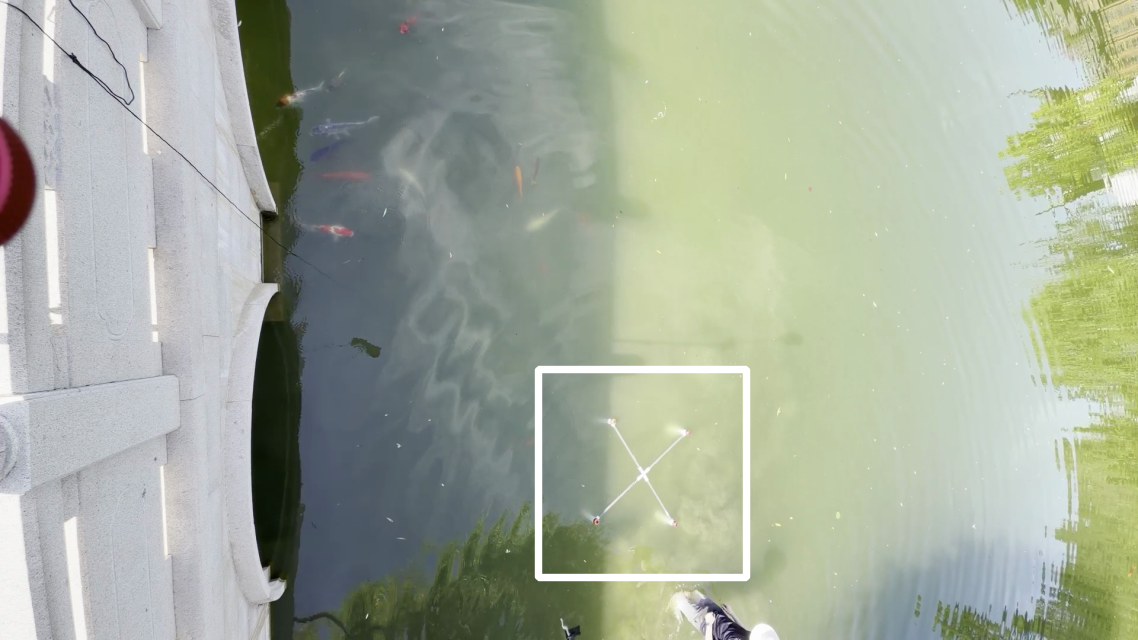}
     \caption{15H: 15-foot (4.57m) setup.}
 \end{subfigure}

 \vspace{0.5em}

 \begin{subfigure}[t]{\textwidth}
     \centering
     \includegraphics[height=26mm]{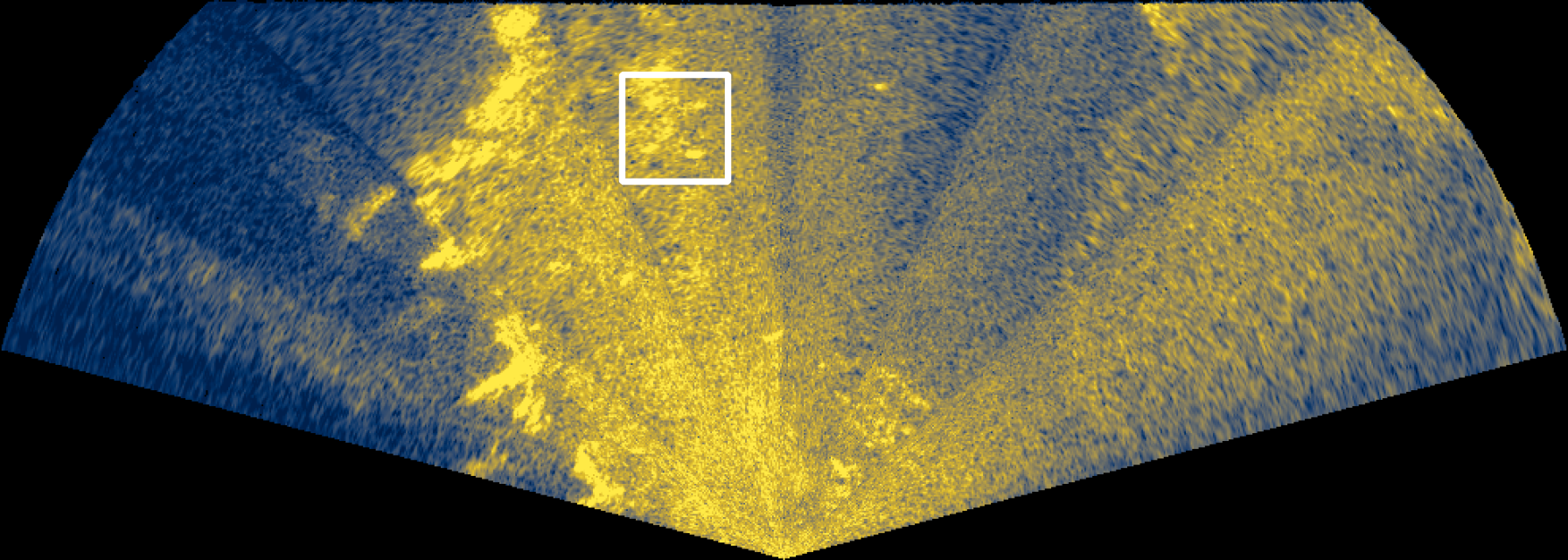}
     \hfill
     \includegraphics[height=26mm]{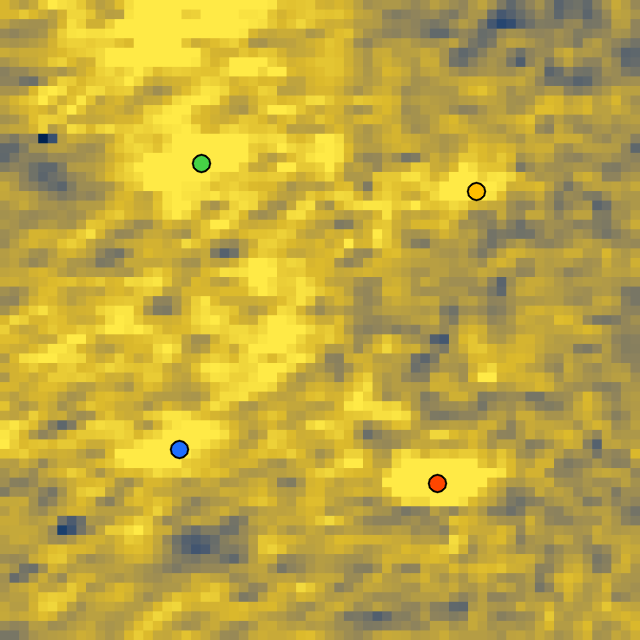}
     \hfill
     \includegraphics[height=26mm]{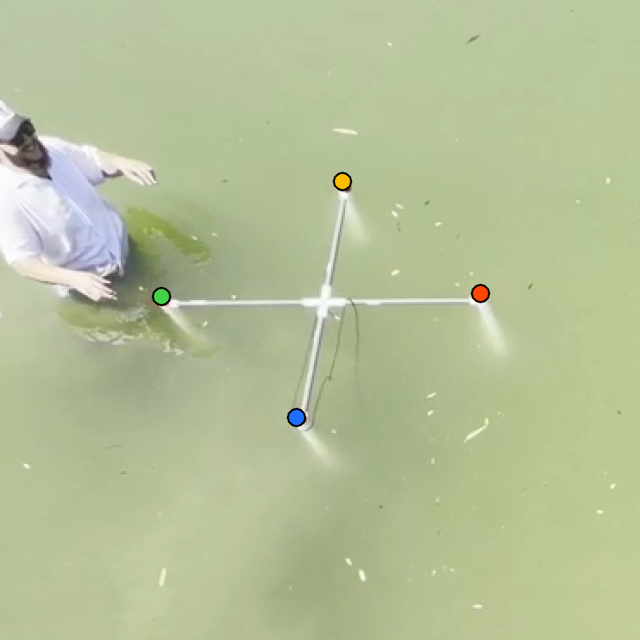}
     \hfill
     \includegraphics[height=26mm]{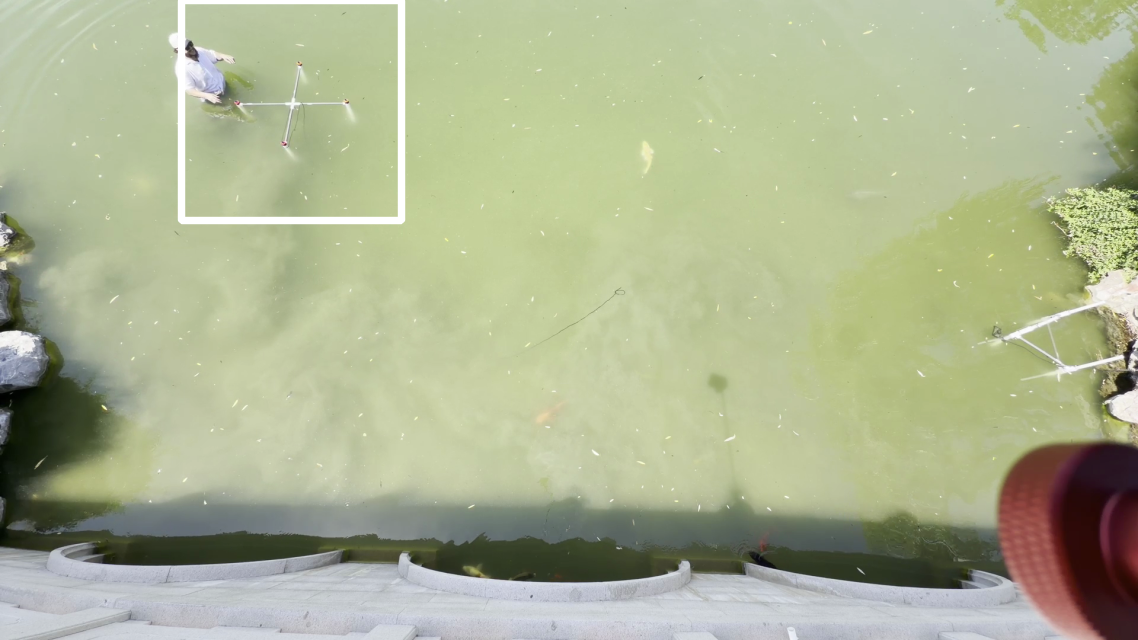}
     \caption{40H: 40-foot (12.19m) setup.}
 \end{subfigure}

 \vspace{0.5em}

 \begin{subfigure}[t]{\textwidth}
     \centering
     \includegraphics[height=26mm]{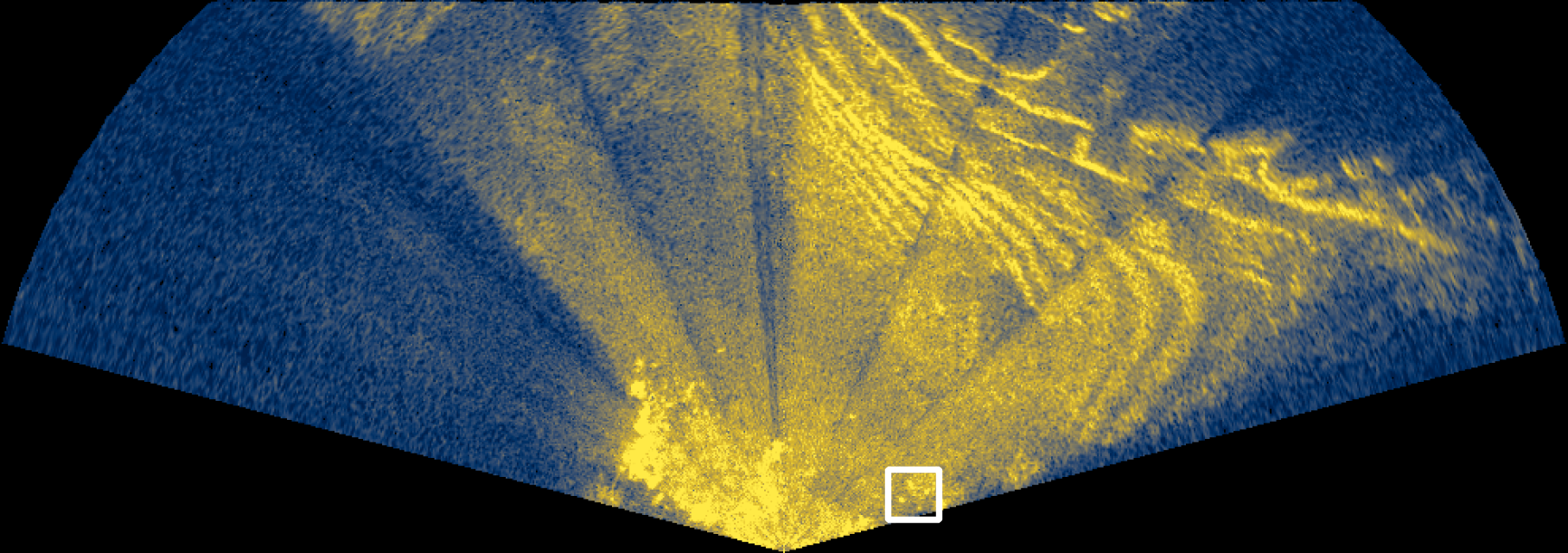}
     \hfill
     \includegraphics[height=26mm]{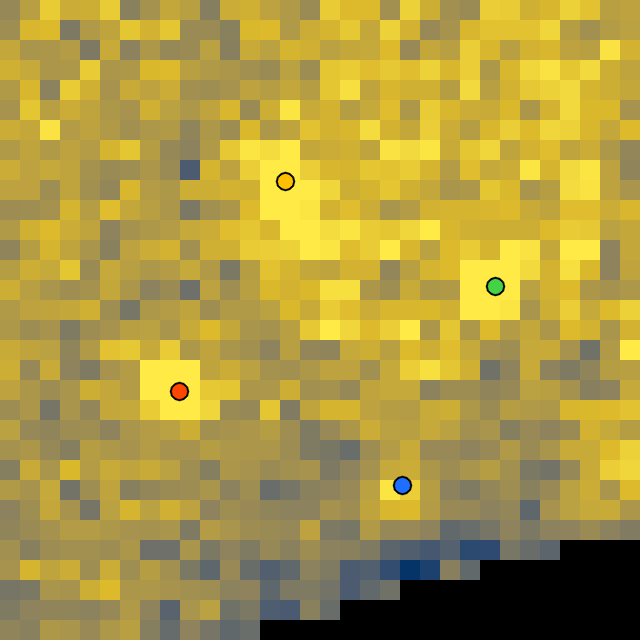}
     \hfill
     \includegraphics[height=26mm]{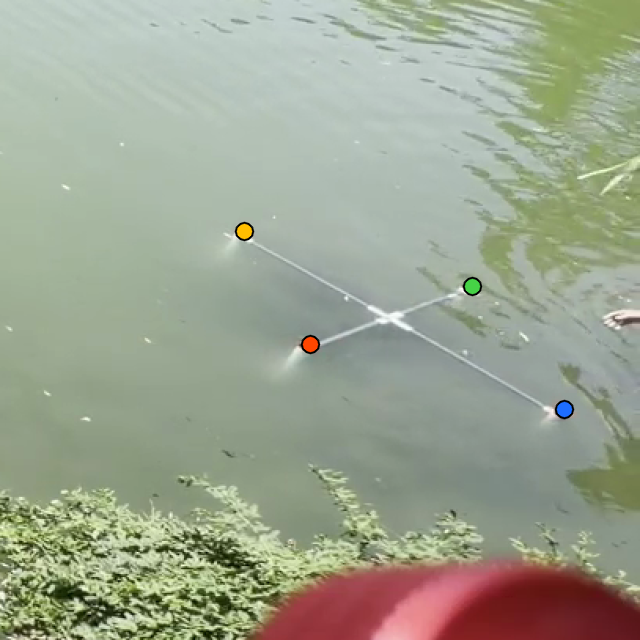}
     \hfill
     \includegraphics[height=26mm]{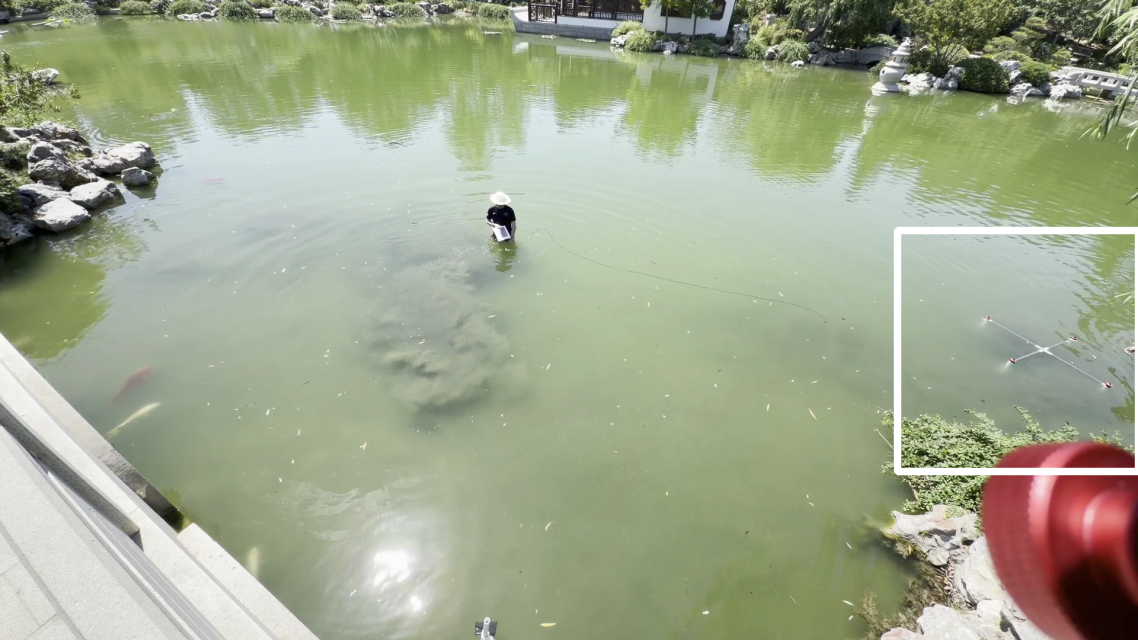}
     \caption{100H: 100-foot (30.48) setup.}
 \end{subfigure}

 \caption{
     Example sonar-RGB correspondences used for homography estimation.
     From left to right: full cartesian sonar frame, cropped sonar region containing the calibration object with homography points,  cropped RGB region containing the calibration object with homography points, and full RGB frame.}
 \label{fig:homography}
\end{figure*}

To estimate spatial relationships between the two imaging modalities we performed planar registration.
To aid in finding homographic correspondences we used a calibration object.
The calibration object is a cross with pipes extending downwards; see the third column of \cref{fig:homography}.
The vertical pipes that descend into the water column are approximately 1.5~m center-to-center and 48~mm in diameter.
The geometric layout of the reference object provided multiple points per frame and known straight lines.
The regular geometry, as opposed to multiple free-floating buoys, also aided manual localization of the object in the noisy sonar imagery.

The tops of the pipes were used as reference points for the RGB imagery and the center of the cross-section of the pipes was used as reference points in the sonar imagery.
This introduces parallax error as the fish are below the surface and the RGB reference points are above the water ($<$25mm).

Corresponding points were manually localized in temporally-matched cartesian-transformed sonar and RGB frames, see \cref{tab:homography_points} for the number of points per run.

\begin{figure*}
 \centering

 \begin{subfigure}[t]{\textwidth}
     \centering
     \includegraphics[width=0.325\textwidth]{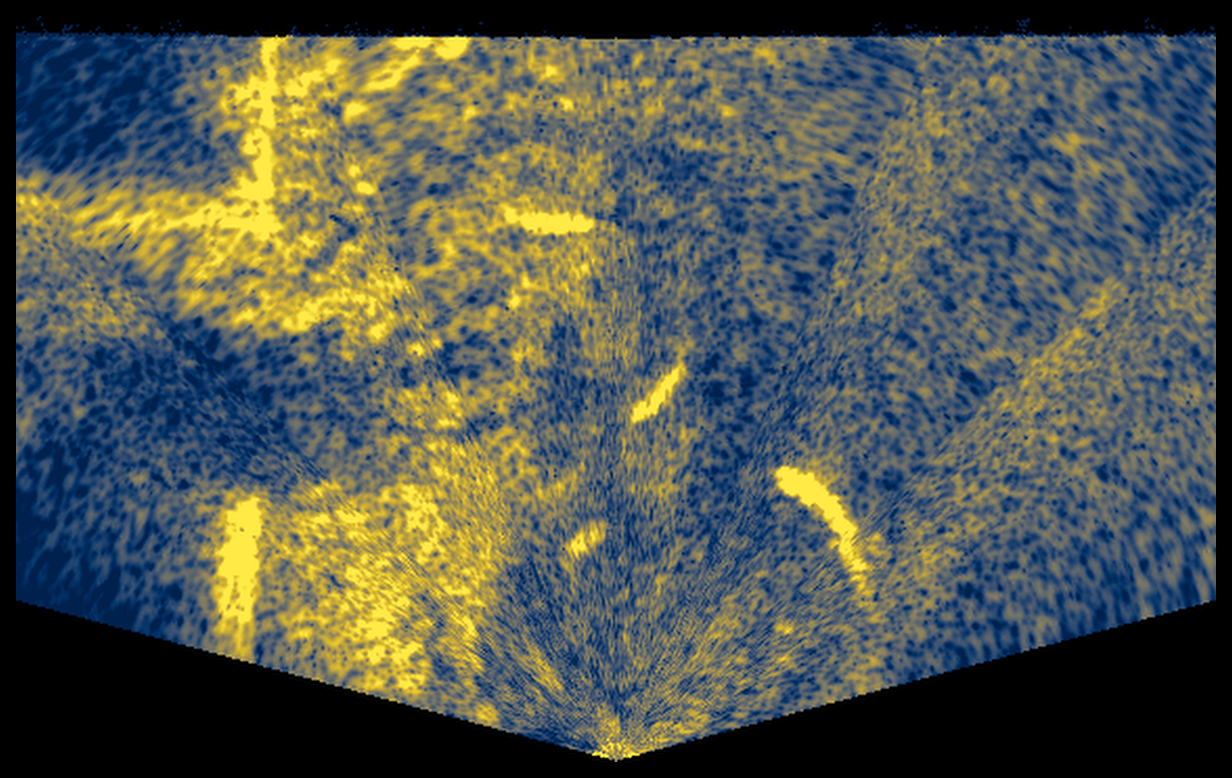}
     \hfill
     \includegraphics[width=0.325\textwidth]{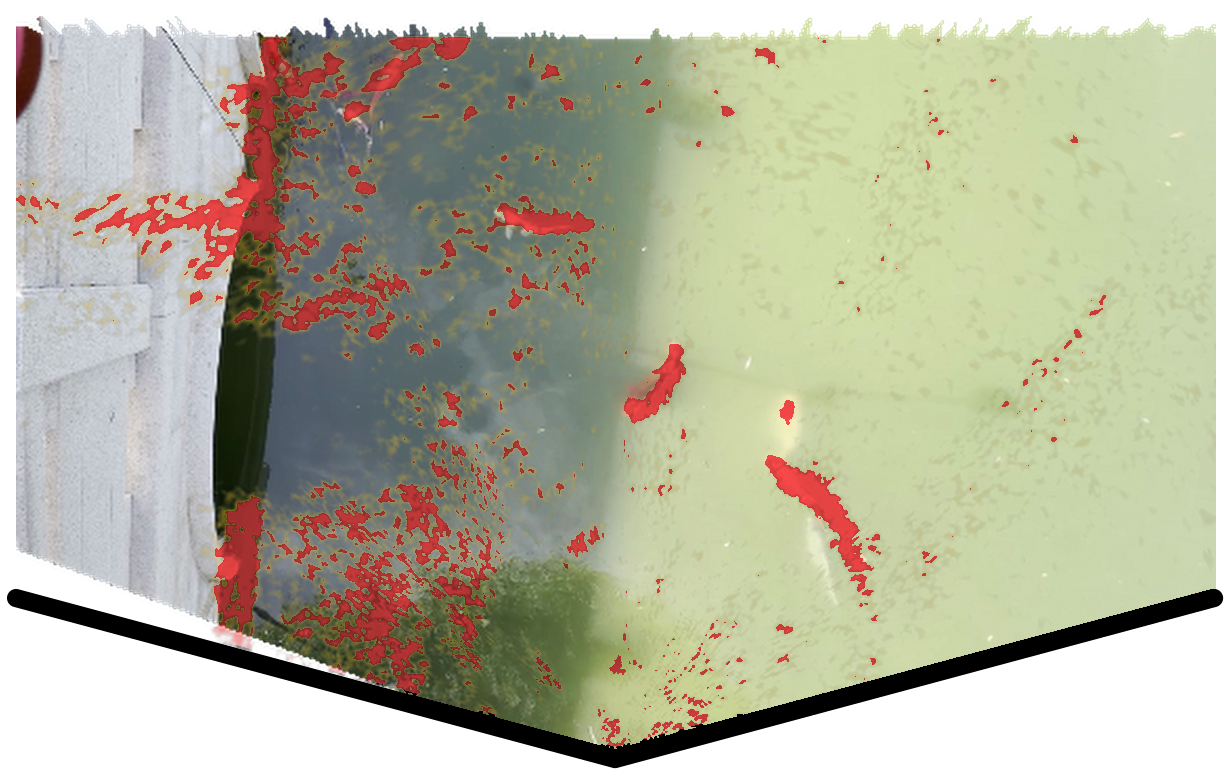}
     \hfill
     \includegraphics[width=0.325\textwidth]{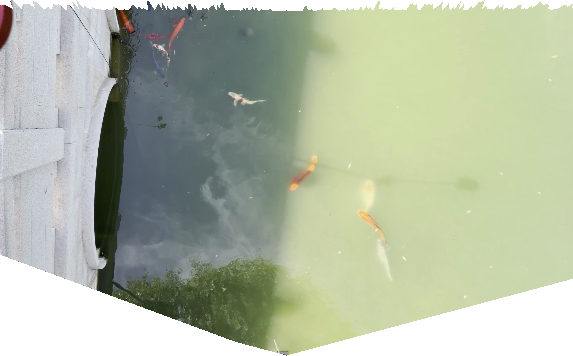}
     \caption{15H: 15-foot (4.57m) homographically aligned imagery (cropped).}
 \end{subfigure}

 \vspace{0.5em}

 \begin{subfigure}[t]{\textwidth}
     \centering
     \includegraphics[width=0.325\textwidth]{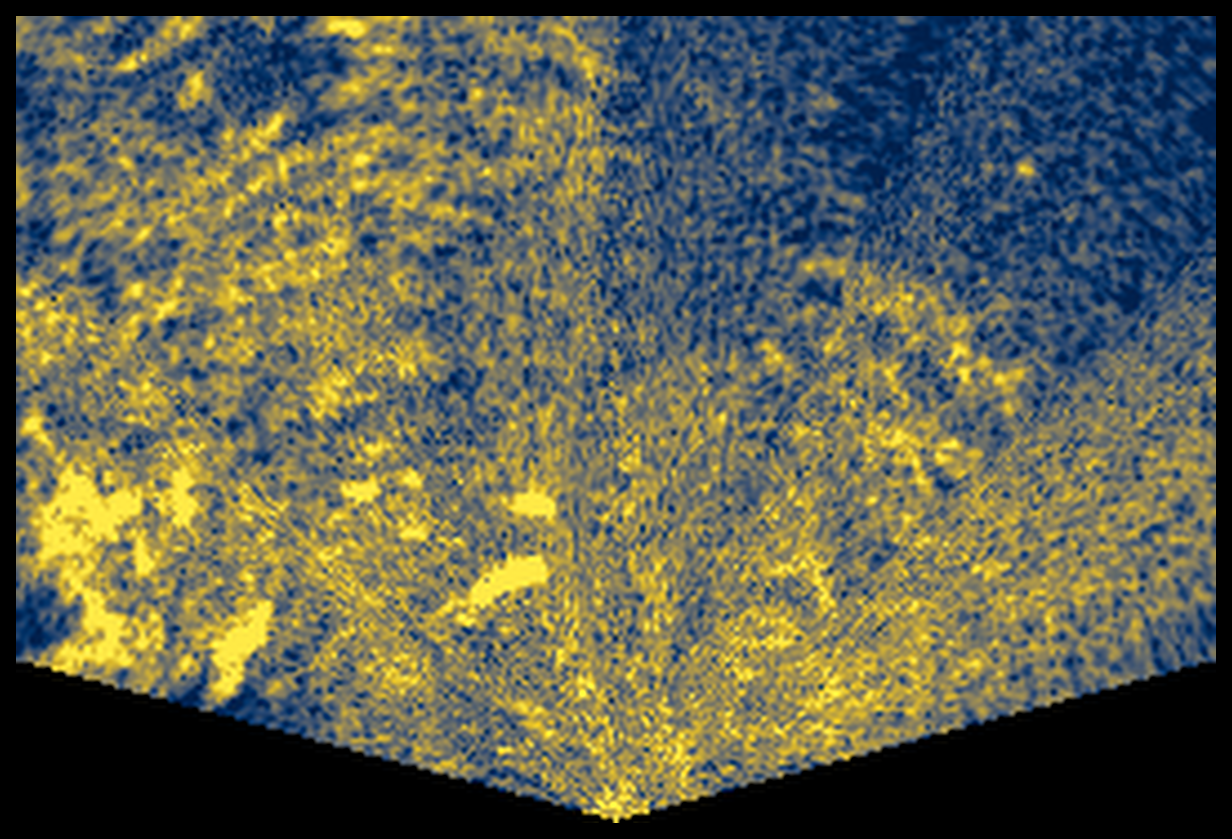}
     \hfill
     \includegraphics[width=0.325\textwidth]{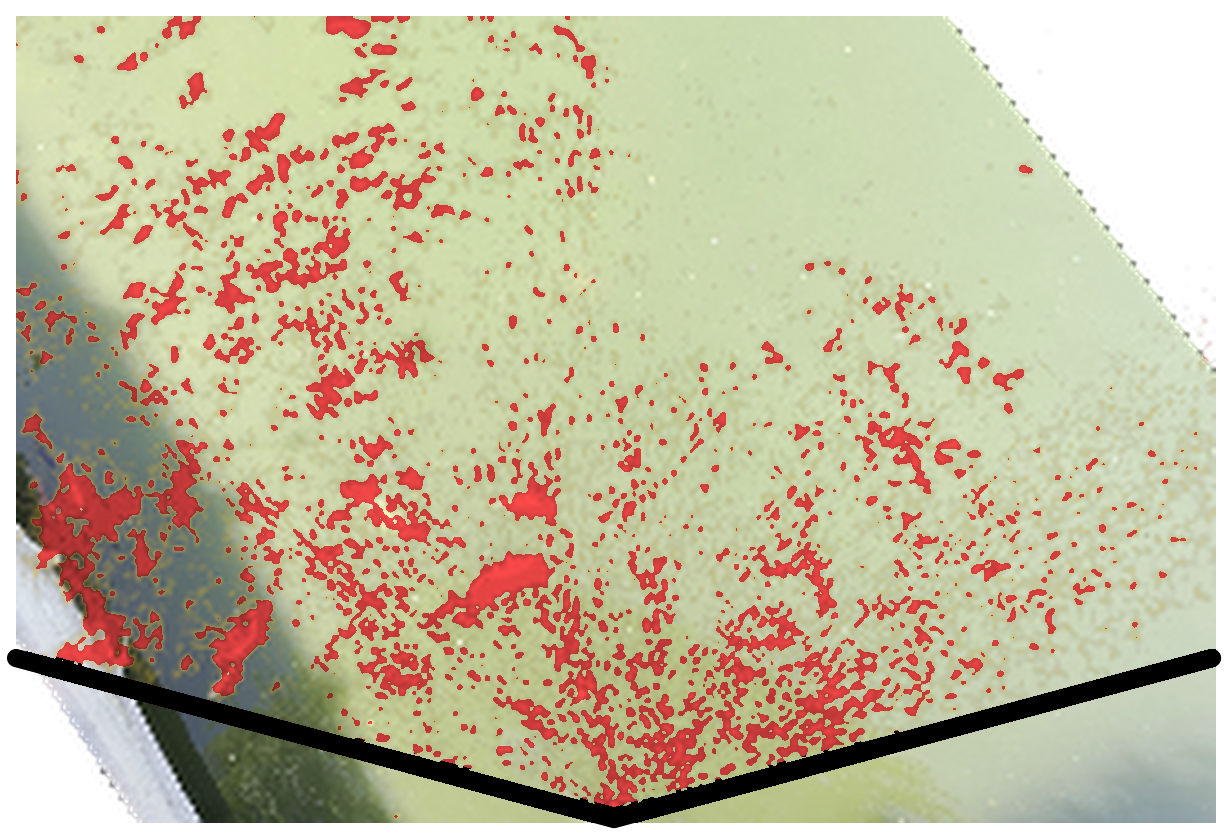}
     \hfill
     \includegraphics[width=0.325\textwidth]{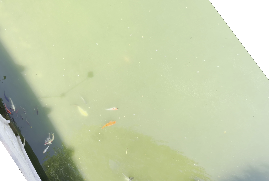}
     \caption{40H: 40-foot (12.19m)  homographically aligned imagery (cropped).}
 \end{subfigure}

 \vspace{0.5em}

 \begin{subfigure}[t]{\textwidth}
     \centering
     \includegraphics[width=0.325\textwidth]{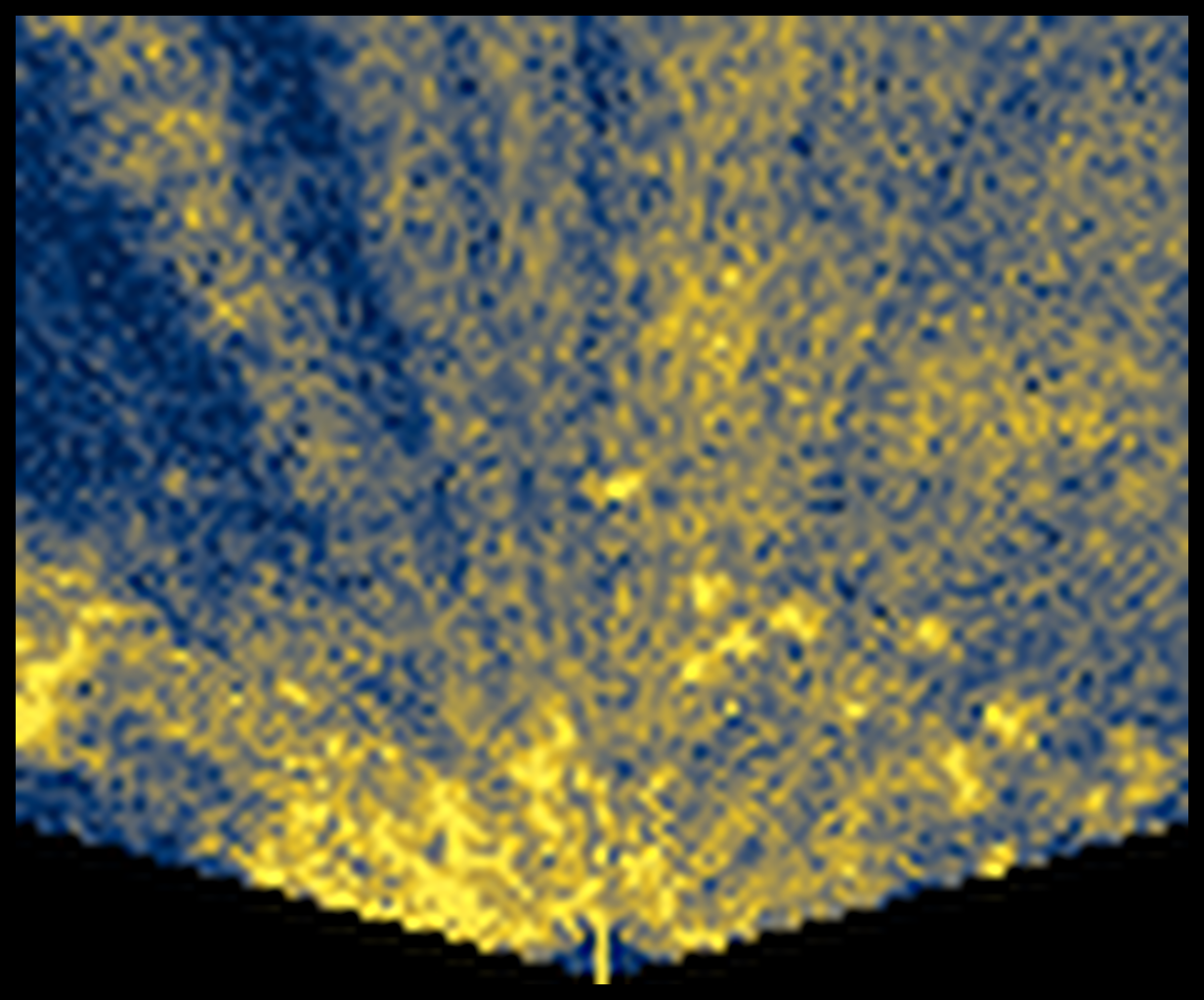}
     \hfill
     \includegraphics[width=0.325\textwidth]{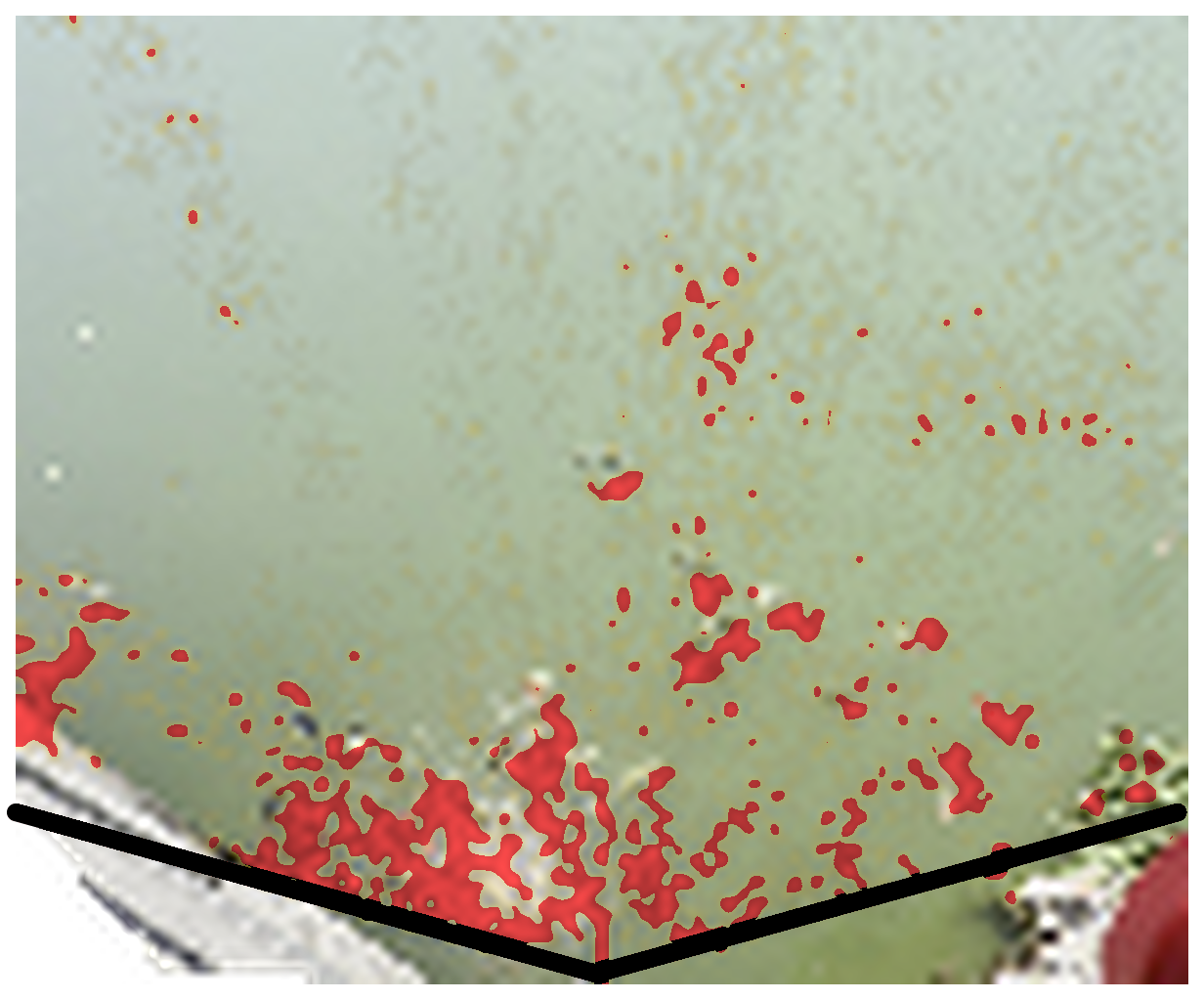}
     \hfill
     \includegraphics[width=0.325\textwidth]{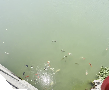}
     \caption{100H: 100-foot (30.48m)  homographically aligned imagery (cropped).}
 \end{subfigure}

 \caption{After homographic warping, the two modalities are approximately aligned.
The RGB frame is homographically warped (right) to align with the sonar frame (left).
As discussed in \cref{sec:homography}, a planar mapping cannot fully capture depth-dependent displacement.}
 \label{fig:post_homography}
\end{figure*}

Homographic mappings were estimated using RANSAC with a 12-RGB-pixel threshold and refined using the consensus points.
The inlier rates were 92.9\%, 87.5\%, and 91.7\% for 15H, 40H, and 100H, respectively.
The inlier RMSE was 2.15 cm, 3.57 cm, and 5.89 cm for 15H, 40H, and 100H, respectively (see \cref{tab:homography_points} for details). Zoomed-in examples after alignment are shown in \cref{fig:post_homography}.

We recognize that there are some errors in this pipeline.
Depth-dependent parallax is one likely source of residual misalignment, which methods using this dataset may need to account for.

Details, including the homography matrices, are provided in \cref{sec:appendix_homogrpahy}; evaluation of the homography accuracy is provided in \cref{sec:appendix_homography_accuracy}.

Frames that included the homography calibration object are not included in the released temporally matched dataset, but are included as additional data.

\begin{figure*}
 \centering
 \begin{subfigure}[t]{0.32\textwidth}
     \centering
     \includegraphics[width=\linewidth]{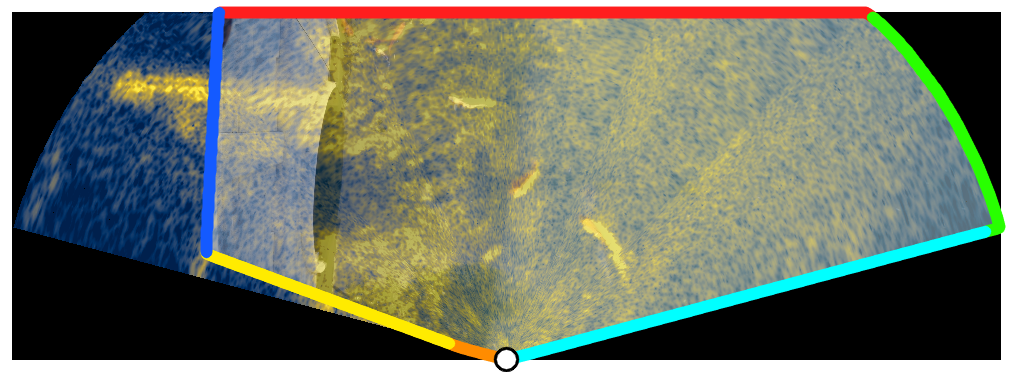}
     \caption{15H sonar, 84.3\% RGB converage}
 \end{subfigure}%
 \hfill 
 \begin{subfigure}[t]{0.32\textwidth}
     \centering
     \includegraphics[width=\linewidth]{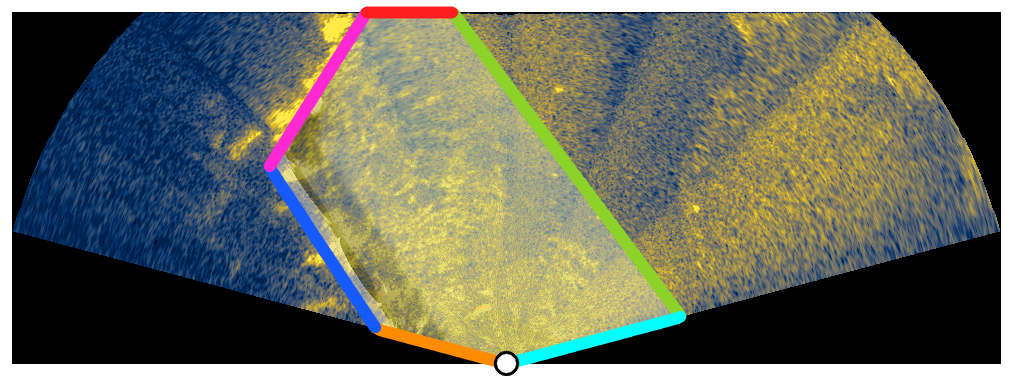}
     \caption{40H sonar, 32.3\% RGB converage}
 \end{subfigure}%
 \hfill 
 \begin{subfigure}[t]{0.32\textwidth}
     \centering
     \includegraphics[width=\linewidth]{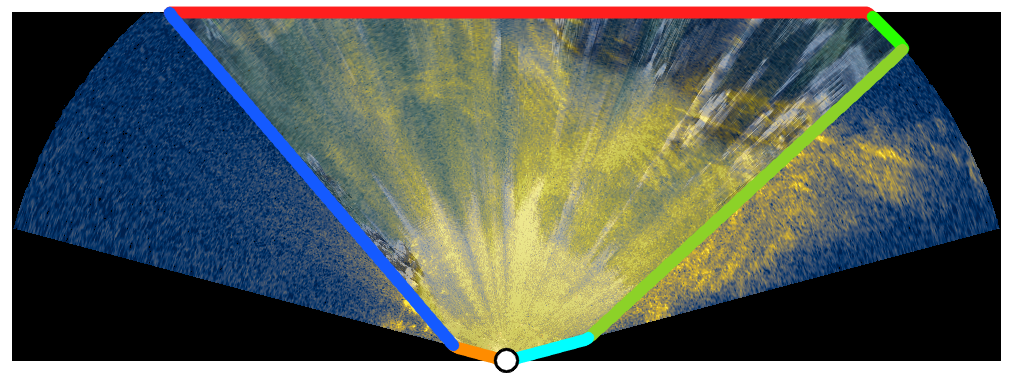}
     \caption{100H sonar, 59.7\% RGB converage}
 \end{subfigure}
 \\
 \begin{subfigure}[t]{0.32\textwidth}
     \centering
     \includegraphics[width=\linewidth]{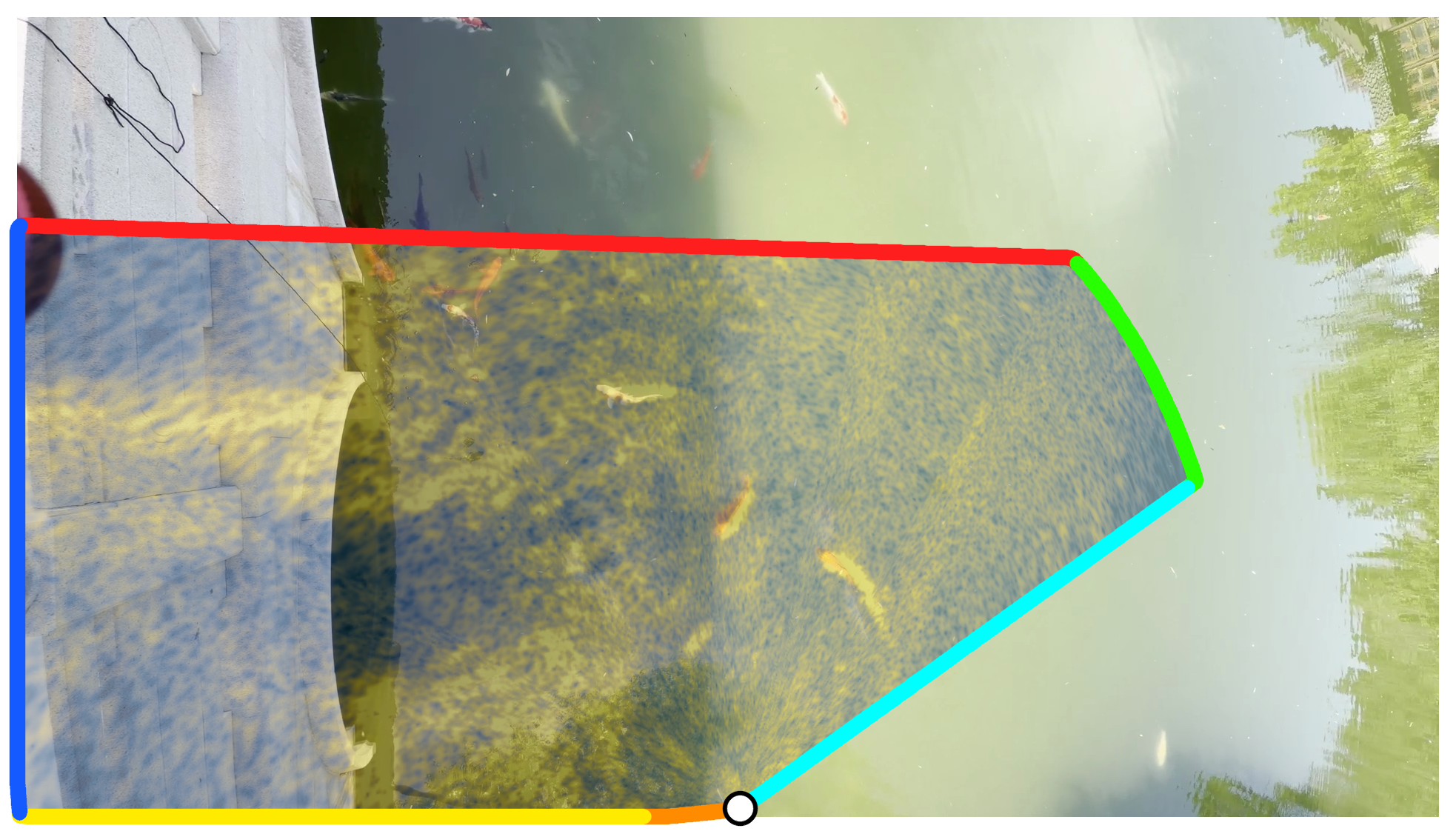}
     \caption{15H RGB, 51.5\% Sonar Coverage}
 \end{subfigure}%
 \hfill 
 \begin{subfigure}[t]{0.32\textwidth}
     \centering
     \includegraphics[width=\linewidth]{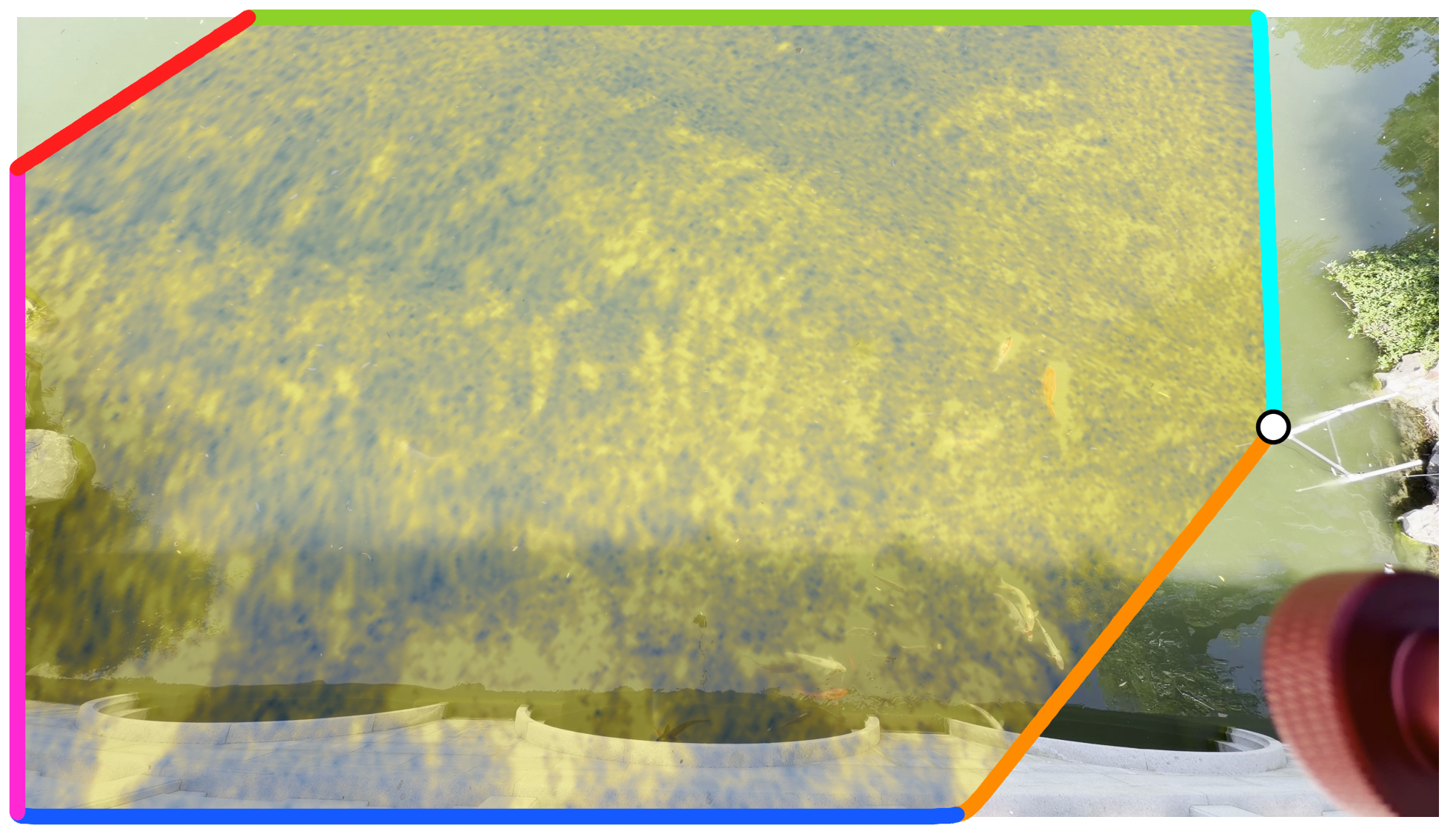}
     \caption{40H RGB, 81.2\% Sonar Coverage}
 \end{subfigure}%
 \hfill 
 \begin{subfigure}[t]{0.32\textwidth}
     \centering
     \includegraphics[width=\linewidth]{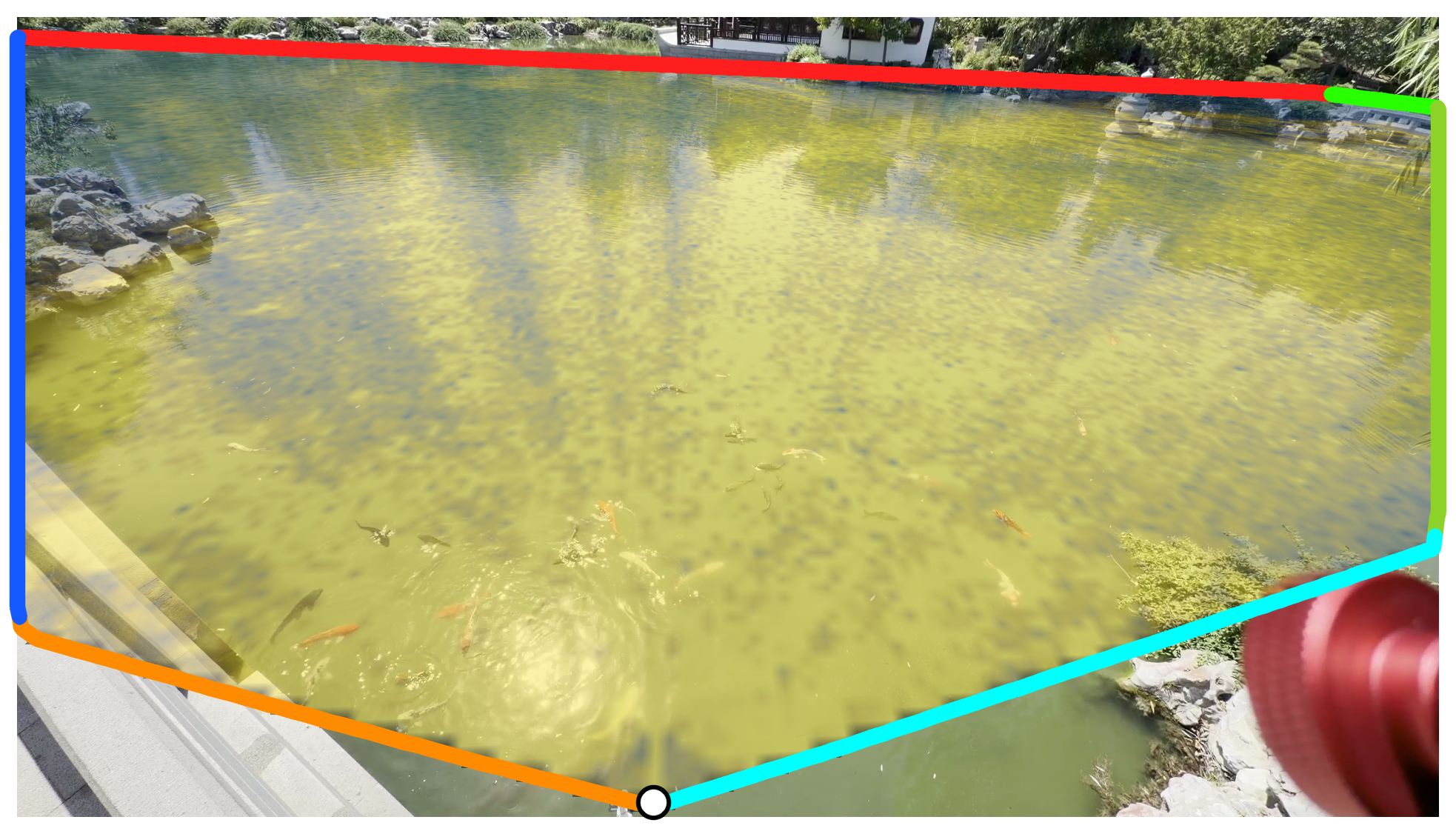}
     \caption{100H RGB, 73.6\% Sonar Coverage}
 \end{subfigure}
 \caption{
 Spatial overlap between sonar and RGB imagery for the three horizontal configurations (15H, 40H, and 100H, left to right), estimated using the calibrated homographies.
Top: RGB coverage in cartesian sonar coordinates, retaining the full recorded sonar footprint.
Bottom: sonar coverage in RGB coordinates.
 Percentages indicate the fraction of each reference image covered by the other modality.}
 \label{fig:homography_overlap}
\end{figure*}

Median cm and RMSE cm values are measured in the sonar plane and represent the physical error between the RGB points projected onto the corresponding sonar points, after converting the resulting sonar-pixel distances to centimeters using the sonar range-bin scale.
These values are specified in \cref{tab:homography_points}.
The resulting spatial overlap between modalities for the horizontal configurations is shown in \cref{fig:homography_overlap}.

\section{Data Availability and Release}
\label{sec:data_availability}

The \name dataset is publicly available on
\href{https://huggingface.co/datasets/perona-lab/Huntington_and_Caltech_Koi_RGB_and_Sonar}{Hugging Face}.
The dataset comprises approximately 229\,GB of data and is distributed under the MIT License.

Documentation describing the dataset organization, metadata fields, coordinate conventions, and example procedures for loading and aligning paired observations is provided alongside the dataset.
Code and tools for working with the dataset are available on
\href{https://github.com/fisheye-sonar/Huntington-and-Caltech-Koi-RGB-and-Sonar-Dataset}{GitHub}.

The sonar data was recorded with \href{https://github.com/mahobley/OpenScope}{OpenScope}, our open-source desktop sonar viewer and recorder for the Garmin GLS 10 and LVS32, which runs on macOS, Windows, and Linux.
OpenScope is available on GitHub under the MIT License and can be used to record additional LiveScope sonar data together with frame-level metadata.

\section{Limitations}
\label{sec:limitations}

\paragraph{Planar geometry and calibration.}
RGB and sonar have different imaging geometries, and the homographies for 15H, 40H, and 100H approximate the relationship between the modalities using a single plane.
Fish, vegetation, and background structures occupy different depths, and the RGB and sonar calibration reference points are not physically coincident.
Refraction at the air--water interface and the finite vertical extent of the sonar beam introduce additional differences between the modalities.
Consequently, planar registration does not recover three-dimensional fish positions, and agreement on the calibration object does not imply equivalent alignment for fish at other depths.
The reported calibration residuals are also evaluated on correspondences involved in model estimation and therefore measure internal calibration agreement rather than independent localization accuracy.
A future collection could improve this evaluation using held-out calibration targets and reference points distributed across multiple positions and depths.
The registration procedure and calibration residuals are described in \cref{sec:homography,sec:appendix_homography_accuracy}.

\paragraph{Temporal alignment.}
The camera and sonar were not synchronized by a shared hardware trigger.
Pairing instead uses corrected device timestamps, supported by physical synchronization events visible in both modalities.
RGB timestamps were mapped to wall-clock time using robust affine fits to the integer-second transitions in the Mac receipt timestamps, then adjusted using median sonar-RGB offsets from manually matched pipe events; the 15-ft recording required separate offsets before and after a timestamp discontinuity.
Capture starts and ends were checked against the frame counts within partial wall-clock seconds, and elapsed recording durations
and inter-capture gaps were evaluated using the recorded timestamps
rather than an assumed constant frame rate.
At the fitted synchronization anchors, the residual timestamp RMSE was 6.0, 6.8, 6.1, and 16.9~ms for the 15-ft, 40-ft, 100-ft, and vertical-plane runs, respectively, with maximum absolute residuals of 6.3, 17.5, 11.0, and 37.3~ms across 4, 8, 6, and 13 annotated pairs.
These are residuals at the events used to estimate the corrections, rather than independent measurements of exposure-time synchronization or guarantees of alignment throughout each recording.
The 100~ms matching threshold bounds the difference between the assigned timestamps, not necessarily the physical acquisition times.
Residual transport latency, dropped or repeated RGB frames, capture-boundary discontinuities, and variable sonar frame rates may remain.
Motion-sensitive analyses should therefore use the supplied timestamps rather than assume uniformly spaced or exposure-synchronized frame pairs.
A future acquisition would benefit from a common hardware timing signal or an independently recorded synchronization source where supported by the devices.
Further details of the timing correction and synchronization quality are provided in \cref{sec:temporal_alignment,sec:appendix_time_sync}.

\paragraph{Incomplete overlap and unequal visibility.}
The two modalities have different fields of view and visibility failure modes, observe only partially overlapping regions, and geometric overlap does not imply equal visibility.
Fish may lie outside the sonar beam's vertical extent while remaining visible to the top-down camera, or may be obscured in RGB by water color, turbidity, or surface reflections.
Conversely, sonar returns may be ambiguous or overlap with environmental structure.
RGB observations should therefore be treated as complementary evidence rather than exhaustive ground truth for individual sonar returns.
Absence from one modality should not automatically be interpreted as absence from the scene.
Cross-modal annotation and evaluation procedures should distinguish between verified negatives and regions or objects that are unobserved or ambiguous.
The measured spatial overlap for the three horizontal configurations is shown in \cref{fig:homography_overlap}.

\paragraph{Range and spatial representation.}
The physical sampling represented by a sonar pixel depends on acquisition geometry and range.
Transforming the measurements into Cartesian coordinates provides a convenient spatial representation but does not increase the information contained in the original sonar observations.
Apparent object size, boundary precision, and separation between nearby fish should therefore be interpreted with respect to the native range and angular sampling rather than the displayed Cartesian raster alone.
The differences among the 15H, 40H, and 100H recordings provide examples of these range-dependent effects, but because the runs also differ in scene content and calibration observations they should not be interpreted as a controlled experiment isolating range alone.

\paragraph{Dataset scope and downstream evaluation.}
\name contains approximately 3.3 hours of selected footage from a single pond, one fish type, one sonar system, and one RGB camera.
It therefore does not establish generalization to other species, sites, sensors, or environmental conditions, including the highly turbid or dark environments that motivate many applications of imaging sonar.
The current release also does not provide task-specific reference detections, identities, tracks, or counts.
Such annotations and appropriate evaluation protocols are required before drawing conclusions about downstream task performance.
Because neighboring video frames are highly correlated, random frame-level train/test splits could also place nearly identical observations, and potentially the same individual fish, in both sets.
Future benchmarks should use temporally separated evaluation intervals and, where possible, independent sites or populations.

\section{Practical Lessons for Future Collections}
\label{sec:lessons}

One purpose of releasing the supporting synchronization and calibration information is to make the collection procedure itself inspectable.
Several aspects of the present dataset suggest straightforward improvements for future RGB-sonar collections.

The most important improvement would be stronger synchronization.
A shared hardware trigger or independently recorded timing signal would reduce uncertainty introduced by separate device clocks, file boundaries, and transport latency.
When hardware synchronization is unavailable, repeated physical synchronization events throughout a recording can help reveal timing drift or discontinuities rather than providing only a single alignment check.
The temporal-alignment procedure and its remaining uncertainties are described in \cref{sec:temporal_alignment,sec:appendix_time_sync}.

Spatial calibration would similarly benefit from being designed around independent validation rather than only model estimation.
Calibration targets could be observed at multiple known depths and positions, with separate observations reserved for evaluating registration accuracy.
For top-down optical imaging, explicitly measuring camera position, orientation, height, and the water-surface geometry would also make it possible to investigate models that account for refraction and three-dimensional scene structure rather than relying entirely on a planar homography.
The planar-registration procedure and its evaluation are described in \cref{sec:homography,sec:appendix_homography_accuracy}.

Finally, field-of-view overlap should ideally be evaluated before the primary recordings begin.
The substantially different footprints produced by different sonar ranges mean that a camera placement appropriate for one configuration may provide limited coverage in another.
A short calibration acquisition at each intended range can be used to verify both geometric overlap and whether fish-sized objects remain sufficiently resolved in each modality.
The resulting spatial overlap for the three horizontal configurations is shown in \cref{fig:homography_overlap}.

These observations are not intended to imply that the present alignment provides exact cross-modal ground truth.
Rather, documenting where uncertainty enters the acquisition pipeline is part of the purpose of the dataset.
We expect that future collections can use these experiences to obtain stronger temporal synchronization, more complete spatial calibration, and greater environmental diversity.

\section{Conclusion}
\label{sec:conclusion}

We introduce \name, a paired dataset of consumer imaging-sonar and top-down RGB recordings of koi carp at The Huntington Library, Art Museum, and Botanical Gardens.
\name includes more than 105,000 temporally matched pairs from four range/orientation configurations spanning 3.3 hours.
In addition, we provide frame-wise metadata, sonar polar-to-Cartesian transforms, supporting synchronization and calibration frames, and derived planar homographies for each of the three horizontal runs.

The paired RGB imagery offers complementary evidence when sonar observations are difficult to interpret and can support annotation, method development, and diagnostic evaluation of sonar-based fish monitoring methods.
However, differences in visibility, depth, and acquisition timing, as well as calibration inaccuracies, mean that the RGB-sonar mapping is not perfectly one-to-one.
Methods utilizing this data should account for these differences in their design.

Within this scope, \name provides a resource for investigating consumer imaging sonar for fish monitoring and for studying how paired optical observations can assist the interpretation and development of sonar-based methods.
Extending the dataset to additional sites, species, and environmental conditions, together with task-specific annotations, would enable more comprehensive evaluation of detection, tracking, and counting methods.

\bibliographystyle{plain} 
\bibliography{bib} 

@article{martignac2015didson,
  author  = {Martignac, Fran\c{c}ois and Daroux, Aur{\'e}lie and Baglini{\`e}re, Jean-Luc and Ombredane, Dominique and Guillard, Jean},
  title   = {The use of acoustic cameras in shallow waters: new hydroacoustic tools for monitoring migratory fish population. A review of {DIDSON} technology},
  journal = {Fish and Fisheries},
  year    = {2014},
  doi     = {10.1111/faf.12071}
}

@article{mccann2018underwater,
  author  = {McCann, Erin L. and Li, Liling and Pangle, Kevin and Johnson, Nicholas S. and Eickholt, Jesse},
  title   = {An underwater observation dataset for fish classification and fishery assessment},
  journal = {Scientific Data},
  year    = {2018},
  doi     = {10.1038/sdata.2018.190}
}

@article{terayama2019integration,
  author  = {Terayama, Kei and Shin, Kento and Mizuno, Katsunori and Tsuda, Koji},
  title   = {Integration of sonar and optical camera images using deep neural network for fish monitoring},
  journal = {Aquacultural Engineering},
  year    = {2019},
  doi     = {10.1016/j.aquaeng.2019.102000}
}

@article{wei2022monitoring,
  author  = {Wei, Yaoguang and Duan, Yunhong and An, Dong},
  title   = {Monitoring fish using imaging sonar: Capacity, challenges and future perspective},
  journal = {Fish and Fisheries},
  year    = {2022},
  doi     = {10.1111/faf.12693}
}

@article{connolly2023shadows,
  author  = {Connolly, R. M. and Jinks, K. I. and Shand, A. and Taylor, M. D. and Gaston, T. F. and Becker, A. and Jinks, E. L.},
  title   = {Out of the shadows: automatic fish detection from acoustic cameras},
  journal = {Aquatic Ecology},
  year    = {2023},
  doi     = {10.1007/s10452-022-09967-5}
}

@article{sibley2023sound,
  author  = {Sibley, Edward C. P. and Elsdon, Travis S. and Marnane, Michael J. and Madgett, Alethea S. and Harvey, Euan S. and Cornulier, Thomas and Driessen, Damon and Fernandes, Paul G.},
  title   = {Sound sees more: A comparison of imaging sonars and optical cameras for estimating fish densities at artificial reefs},
  journal = {Fisheries Research},
  year    = {2023},
  doi     = {10.1016/j.fishres.2023.106720}
}

@article{duan2023boosting,
  author  = {Duan, Yunhong and Zhang, Shubin and Liu, Yang and Liu, Jincun and An, Dong and Wei, Yaoguang},
  title   = {Boosting fish counting in sonar images with global attention and point supervision},
  journal = {Engineering Applications of Artificial Intelligence},
  year    = {2023},
  doi     = {10.1016/j.engappai.2023.107093}
}

@inproceedings{dahn2024acoustic,
  author    = {Dahn, Nikolas and Bande Firvida, Miguel and Sharma, Proneet Kumar and Christensen, Leif and Geisler, Oliver and Mohrmann, Jochen and Frey, Torsten and Sanghamreddy, Prithvi and Kirchner, Frank},
  title     = {An Acoustic and Optical Dataset for the Perception of Underwater Unexploded Ordnance ({UXO})},
  booktitle = {OCEANS 2024},
  year      = {2024},
  publisher = {IEEE},
  doi       = {10.1109/OCEANS55160.2024.10754316}
}

@article{li2025rgbsonar,
  author        = {Li, Yunfeng and Wang, Bo and Sun, Jiuran and Wu, Xueyi and Li, Ye},
  title         = {{RGB-Sonar} Tracking Benchmark and Spatial Cross-Attention Transformer Tracker},
  journal       = {IEEE Transactions on Circuits and Systems for Video Technology},
  year          = {2025},
  doi           = {10.1109/TCSVT.2024.3497214},
  eprint        = {2406.07189},
  archivePrefix = {arXiv},
  url           = {https://arxiv.org/abs/2406.07189}
}

@ARTICLE{groger2024observatory,
    
AUTHOR={Gröger, Joachim P.  and Cisewski, Boris  and Badri-Hoeher, Sabah  and Böer, Gordon  and Boos, Karin  and Clemmesen, Catriona  and Cojocaru, Ala  and Dauben, Verena  and Hoeher, Peter A.  and Lehmann, Andreas  and Matz, Sebastian  and Mehrtens, Hela  and Mittermayer, Felix  and Renkewitz, Helge  and Schramm, Hauke  and Strickmann, Tobias  and Westphalen, Jonni  and Wilts, Thomas  and Winkler, Julian  and Wolf, Dennis  and Zenk, Oliver },

TITLE={Development and operation of a novel non-invasive opto-acoustic underwater fish observatory in Kiel Bight, Southwestern Baltic Sea},
          
JOURNAL={Frontiers in Marine Science},
           
YEAR={2024},
  
URL={https://www.frontiersin.org/journals/marine-science/articles/10.3389/fmars.2024.1425259},
  
DOI={10.3389/fmars.2024.1425259},
  
ISSN={2296-7745},
}

@article{wilts2025hodor,
  author  = {Wilts, Thomas and B{\"o}er, Gordon and Winkler, Julian and Cisewski, Boris and Schramm, Hauke and Badri-Hoeher, Sabah},
  title   = {Descriptor: Hydroacoustic and Optical Dataset for Oceanic Research ({HODOR})},
  journal = {IEEE Data Descriptions},
  year    = {2025},
  doi     = {10.1109/IEEEDATA.2025.3596913},
  note    = {Associated dataset DOI: 10.1594/PANGAEA.980000}
}

@misc{ohrem2025solaqua,
  author        = {Ohrem, Sveinung Johan and Haugal{\o}kken, Bent and Kelasidi, Eleni},
  title         = {{SOLAQUA}: {SINTEF} Ocean Large Aquaculture Robotics Dataset},
  year          = {2025},
  eprint        = {2504.01790},
  archivePrefix = {arXiv},
  doi           = {10.48550/arXiv.2504.01790},
  url           = {https://arxiv.org/abs/2504.01790}
}

@article{shen2024didson,
  author  = {Shen, Wei and Liu, Mengqi and Lu, Quanshui and Yin, Zhaowei and Zhang, Jin},
  title   = {A Fish Target Identification and Counting Method Based on {DIDSON} Sonar and {YOLOv5} Model},
  journal = {Fishes},
  year    = {2024},
  doi     = {10.3390/fishes9090346}
}

@article{munnelly2024applications,
  author  = {Munnelly, Ryan T. and Castillo, Jose C. and Handegard, Nils Olav and Kimball, Matthew E. and Boswell, Kevin M. and Rieucau, Guillaume},
  title   = {Applications and analytical approaches using imaging sonar for quantifying behavioural interactions among aquatic organisms and their environment},
  journal = {ICES Journal of Marine Science},
  year    = {2024},
  doi     = {10.1093/icesjms/fsad182}
}

@article{dickson2026sonar,
  author={Dickson, Jon and Watson, Maryann and Dye, Bass and Neefjes, Madelon and van Dorp, Roos and Jasker, Anne and Franken, Oscar and Bouma, Tjeerd and Govers, Laura and Olff, Han and others},
  title   = {Sonar, so good: Tree-reefs drive net gain in fish size, abundance, and diversity},
  journal = {Marine Ecology Progress Series},
  year    = {2026},
  doi     = {10.3354/meps15048}
}

@misc{chen2026sovis,
  author        = {Chen, Weitung and Tinn, Phil and Auran, Per Gunnar and Ludvigsen, Martin and Haro, Peter Halland},
  title         = {A Sonar-Visual Dataset for Cross-Modal Underwater Robot Perception},
  year          = {2026},
  eprint        = {2606.01398},
  archivePrefix = {arXiv},
  doi           = {10.48550/arXiv.2606.01398},
  url           = {https://arxiv.org/abs/2606.01398}
}

@inproceedings{cfc2026neurips,
  title     = {{CFC26}: Building Evaluations for Deployment in Sonar-Based Fish Counting},
  author    = {Van Horn, Madison and Stathatos, Suzanne and Brodjian, Sevan and Kay, Justin and Van Brunt, Kai and Young, Erik and Beery, Sara and Perona, Pietro and Hobley, Michael},
  booktitle = {Advances in Neural Information Processing Systems (NeurIPS), Evaluations \& Datasets Track},
  year      = {2026}
}

@inproceedings{cfc2022eccv,
  author    = {Kay, Justin and Kulits, Peter and Stathatos, Suzanne and Deng, Siqi and Young, Erik and Beery, Sara and Van Horn, Grant and Perona, Pietro},
  title     = {The Caltech Fish Counting Dataset: A Benchmark for Multiple-Object Tracking and Counting},
  booktitle = {Proceedings of the European Conference on Computer Vision (ECCV)},
  year      = {2022},
  doi       = {10.1007/978-3-031-20074-8_17}
}

@inproceedings{stathatos2026saved,
  title     = {{SAVeD}: Learning to Denoise Low-SNR Video for Improved Downstream Performance},
  author    = {Stathatos, Suzanne and Hobley, Michael and Perona, Pietro and Marks, Markus},
  booktitle = {Proceedings of the IEEE/CVF Winter Conference on Applications of Computer Vision (WACV)},
  year      = {2026},
  doi       = {10.1109/WACV61042.2026.00662}
}

@manual{garmin2021livescope,
  author = {{Garmin Ltd.}},
  title  = {\href{https://static.garmin.com/pumac/Panoptix_Livescope_Install_EN.pdf}{Panoptix {LiveScope} Installation Instructions}},
  month  = mar,
  year   = {2021},
  note   = {LVS32 specifications}
}

\newpage
\clearpage
\appendix

\section{Comparison of Related Datasets and Studies}
The main text discusses prior work on imaging sonar for fish monitoring, paired optical--acoustic observations, and multimodal underwater datasets. Here, we provide a structured comparison of the most closely related datasets and studies along several dimensions relevant to \name. \Cref{tab:related_datasets} summarizes whether each resource focuses on fish, includes paired or synchronized optical observations, makes data publicly available, provides spatial calibration between modalities, uses research-grade or consumer imaging sonar, and is intended to support cross-modal learning. The comparison is not intended to be exhaustive, rather, it highlights how existing resources combine these characteristics and provides additional context for the design and intended use of \name.

\begin{table*}[t]
    \centering
    \small
    \renewcommand{\arraystretch}{1.15}

    \setlength{\tabcolsep}{3pt}
    \begin{tabular*}{\textwidth}{
        @{\extracolsep{\fill}}
        l
        c c c c c c
        @{}
    }
        \toprule
        Resource &
        Fish-focused &
        Paired optics &
        Public data &
        Spatial calib. &
        Sonar class &
        \makecell[c]{Cross-modal\\ML intent} \\
        \midrule

        McCann 2018~\cite{mccann2018underwater}
            & \featyes
            & \featyes
            & \featyes
            & \featpartial
            & Research
            & \featpartial \\

        Terayama 2019~\cite{terayama2019integration}
            & \featyes
            & \featyes
            & \featno/\featpartial
            & \featpartial
            & Research
            & \featyes \\

        RGBS50~\cite{li2025rgbsonar}
            & \featno/\featpartial
            & \featyes
            & \featyes
            & \featpartial
            & Research
            & \featyes \\

        DFKI UXO~\cite{dahn2024acoustic}
            & \featno
            & \featyes
            & \featyes
            & \featyes
            & Research
            & \featyes \\

        SOLAQUA~\cite{ohrem2025solaqua}
            & \featyes/\featpartial
            & \featyes
            & \featyes
            & \featpartial
            & Research
            & \featpartial \\

        HODOR~\cite{wilts2025hodor,groger2024observatory}
            & \featyes
            & \featyes
            & \featyes
            & \featpartial
            & Research
            & \featyes \\

        SOVIS~\cite{chen2026sovis}
            & \featyes/\featpartial
            & \featyes
            & \featyes
            & \featpartial
            & Research
            & \featyes \\

        LiveScope ecological study~\cite{dickson2026sonar}
            & \featyes
            & \featno
            & \featno/\featpartial
            & \featno
            & Recreational
            & \featno \\

        \addlinespace[1pt]
                \midrule

        \textbf{\name}
            & \textbf{\featyes}
            & \textbf{\featyes}
            & \textbf{\featyes}
            & \featyes{} 3/4
            & Recreational
            & \textbf{\featyes} \\

        \bottomrule
    \end{tabular*}
    \caption{
        Comparison of relevant multimodal and imaging-sonar datasets and studies.
        \featyes{} indicates that the feature is a central, explicitly documented
        part of the resource; \featpartial{} indicates that it is present but
        secondary, partial, or differently configured; and \featno{} indicates
        that it is absent or not central.
        ``Paired optics'' denotes synchronized optical imagery.
        ``Cross-modal ML intent'' indicates that the resource is explicitly
        framed for fusion, cross-modal learning, translation, or using one
        modality to inform another, rather than merely deploying two sensors
        simultaneously.
    }
    \label{tab:related_datasets}
\end{table*}

\section{Sonar frame inventory}
\Cref{tab:sonar_inventory} lists the capture times, durations, and extracted frame counts for each sonar clip. The 17 listed clips contain 134,220 frames before usable-footage selection and temporal matching: 42,307 in 15H, 39,895 in 40H, 25,830 in 100H, and 26,188 in 20V. Most clips last approximately 15 minutes, with shorter final clips in the horizontal runs.

\section{RGB video inventory}
\Cref{tab:rgb_inventory} summarizes the capture times, durations, and frame counts for the listed RGB clips. These 19 clips contain 479,661 frames, with approximately 27,000 frames in each full 15-minute clip. The RGB and sonar recording intervals do not coincide exactly, in particular, the RGB recording for 40H extends beyond the sonar recording.

\begin{table*}[t]
    \centering
    \scriptsize
    \setlength{\tabcolsep}{3pt}
    \renewcommand{\arraystretch}{1.05}

    \begin{subtable}[t]{0.49\textwidth}
        \centering
        \caption{Sonar frame inventory.}
        \label{tab:sonar_inventory}

        \begin{adjustbox}{max width=\linewidth}
        \begin{tabular}{rrrrrr}
            \toprule
            Run & Clip \# & Capture start & Capture end & Duration (s) & Frames \\
            \midrule
            15H  & 001 & 09:21:23.47 & 09:36:23.99 & 15:00.52 & 13,643 \\
            15H  & 002 & 09:36:24.05 & 09:51:23.90 & 14:59.85 & 10,007 \\
            15H  & 003 & 09:51:24.03 & 10:06:23.87 & 14:59.84 & 7,715  \\
            15H  & 004 & 10:06:24.02 & 10:21:23.90 & 14:59.87 & 6,375  \\
            15H  & 005 & 10:21:24.03 & 10:33:36.93 & 12:12.90 & 4,567  \\
            \midrule
            40H  & 001 & 11:01:43.30 & 11:16:43.98 & 15:00.69 & 9,911  \\
            40H  & 002 & 11:16:44.05 & 11:31:43.78 & 14:59.73 & 10,538 \\
            40H  & 003 & 11:31:43.96 & 11:46:43.96 & 15:00.00 & 6,261  \\
            40H  & 004 & 11:46:44.08 & 12:01:43.92 & 14:59.84 & 5,297  \\
            40H  & 005 & 12:01:44.06 & 12:16:43.89 & 14:59.84 & 4,867  \\
            40H  & 006 & 12:16:44.12 & 12:26:59.19 & 10:15.07 & 3,021  \\
            \midrule
            100H & 001 & 13:47:00.27 & 14:02:00.93 & 15:00.66 & 7,625  \\
            100H & 002 & 14:02:01.17 & 14:17:00.99 & 14:59.82 & 7,322  \\
            100H & 003 & 14:17:01.22 & 14:32:00.73 & 14:59.52 & 6,414  \\
            100H & 004 & 14:32:00.95 & 14:44:56.71 & 12:55.76 & 4,469  \\
            \midrule
            20V  & 001 & 12:56:51.07 & 13:11:51.99 & 15:00.92 & 13,272 \\
            20V  & 002 & 13:11:52.06 & 13:26:51.93 & 14:59.88 & 12,916 \\
            \bottomrule
        \end{tabular}
        \end{adjustbox}
    \end{subtable}
    \hfill
    \begin{subtable}[t]{0.49\textwidth}
        \centering
        \caption{RGB video inventory.}
        \label{tab:rgb_inventory}

        \begin{adjustbox}{max width=\linewidth}
        \begin{tabular}{rrrrrr}
            \toprule
            Run & Clip \# & Capture start & Capture end & Duration (s) & Frames \\
            \midrule
            15H  & 001 & 09:21:35.29 & 09:36:35.19 & 14:59.89 & 26,936 \\
            15H  & 002 & 09:36:36.01 & 09:51:35.24 & 14:59.21 & 26,977 \\
            15H  & 003 & 09:51:36.05 & 10:06:35.28 & 14:59.24 & 26,978 \\
            15H  & 004 & 10:06:36.06 & 10:21:35.27 & 14:59.27 & 26,979 \\
            15H  & 005 & 10:21:36.02 & 10:32:41.68 & 11:05.67 & 19,971 \\
            \midrule
            40H  & 001 & 11:01:45.12 & 11:16:44.38 & 14:59.27 & 26,979 \\
            40H  & 002 & 11:16:45.11 & 11:31:44.44 & 14:59.30 & 26,980 \\
            40H  & 003 & 11:31:45.12 & 11:46:44.45 & 14:59.34 & 26,981 \\
            40H  & 004 & 11:46:45.12 & 12:01:44.48 & 14:59.37 & 26,982 \\
            40H  & 005 & 12:01:45.12 & 12:16:44.52 & 14:59.40 & 26,983 \\
            40H  & 006 & 12:16:45.11 & 12:31:44.59 & 14:59.44 & 26,979 \\
            40H  & 007 & 12:31:45.15 & 12:46:44.62 & 14:59.47 & 26,985 \\
            40H  & 008 & 12:46:45.15 & 12:50:06.98 & 03:21.83 & 6,056  \\
            \midrule
            100H & 001 & 13:46:55.18 & 14:01:54.75 & 14:59.57 & 26,988 \\
            100H & 002 & 14:01:55.16 & 14:16:54.86 & 14:59.60 & 26,989 \\
            100H & 003 & 14:16:55.25 & 14:31:54.88 & 14:59.64 & 26,990 \\
            100H & 004 & 14:31:55.29 & 14:44:07.07 & 12:11.81 & 21,955 \\
            \midrule
            20V  & 001 & 12:56:46.15 & 13:11:45.69 & 14:59.50 & 26,986 \\
            20V  & 002 & 13:11:46.18 & 13:26:45.72 & 14:59.54 & 26,987 \\
            \bottomrule
        \end{tabular}
        \end{adjustbox}
    \end{subtable}

    \caption{
        Recording inventory for the paired sonar and RGB data.
        Each table lists the capture interval, duration, and number of extracted
        frames for each clip and recording configuration.
        For RGB, duration is calculated from the final precise timestamp rather
        than from the nominal frame count divided by 30.
    }
    \label{tab:recording_inventory}
\end{table*}

\section{Framewise metadata}\label{sec:framewise_metadata}
For each sonar-RGB frame pair we have the associated metadata:
\paragraph{Sonar:}
\begin{itemize}
    \item \textbf{Sonar frame ID:} Identifier of the sonar frame within the recorded dataset.
    \item \textbf{Sonar capture timestamp:} Absolute wall-clock acquisition time, including the date and time-zone offset.
    \item \textbf{Sonar clip:} Identifier of the source sonar recording segment containing the frame.
    \item \textbf{Polar bins and angles:} Number of radial range samples and angular beam samples in the polar sonar representation. Each sample records sonar intensity at a particular range and bearing.
    \item \textbf{Range (m):} Maximum radial distance represented by the sonar frame, expressed in meters.
    \item \textbf{$\theta_{\min}$ and $\theta_{\max}$:} Minimum and maximum beam bearings, expressed in degrees relative to the sonar's angular reference axis. Together, they specify the angular extent of the sonar frame.
\end{itemize}
\paragraph{RGB:}
\begin{itemize}
    \item \textbf{RGB frame ID:} Identifier of the RGB frame within the recorded dataset.
    \item \textbf{RGB frame timestamp:} Frame time expressed as
    elapsed seconds relative to the RGB capture's time origin,
    and as an estimated absolute wall-clock time obtained through
    clock calibration. The additional offset used to synchronize
    RGB and sonar is recorded separately.
    \item \textbf{RGB capture number:} Identifier of the source RGB recording containing the frame.
    \item \textbf{$f_x, f_y, c_x, c_y$:} Camera intrinsic parameters for the native, unrotated RGB image. The focal lengths $f_x$ and $f_y$ are expressed in horizontal and vertical pixel units, respectively; $(c_x,c_y)$ gives the principal-point location in image coordinates.
    \item \textbf{Width and height:} Dimensions of the RGB image in pixels.
    \item \textbf{Lens position:} Camera-reported lens focus position at acquisition, rather than a physical distance to the subject.
    \item \textbf{Time received on Mac:} Timestamp recorded when the receiving Mac received the frame data; this is distinct from the acquisition timestamp.
    \item \textbf{Focus, exposure, and white-balance locks:} Boolean indicators specifying whether each camera control was locked during acquisition. A locked control remains fixed rather than being adjusted automatically.
\end{itemize}

\section{Sonar polar range-line position}
\begin{figure*}[htbp]
    \centering
    \newsavebox{\mytable}
    \sbox{\mytable}{%
        \resizebox{0.80\textwidth}{!}{%
            \begin{tabular}{rcccccc}
    \toprule
        Run	&	Min (px)	&	Median (px)	&	Mean (px)	&	Max (px)	&	Std. Dev. (px) 	&	Std. Dev. (\%)	\\
\hline													
15H	&	345	&	345	&	345.06	&	351	&	0.37	&	0.11	\\
40H	&	349	&	349	&	349.17	&	351	&	0.45	&	0.13	\\
100H	&	346	&	346	&	346.97	&	351	&	1.48	&	0.43	\\
20V & 345 & 347 & 347.22 & 351 & 0.90 & 0.26 \\
    \bottomrule
    \end{tabular}
        }%
    }
    \begin{minipage}[t]{0.48\textwidth}
    \vspace{0pt}
    \centering
    \usebox{\mytable}
    \end{minipage}
    \hfill
    \begin{minipage}[t]{0.20\textwidth}
    \vspace{0pt}
    \centering
    \includegraphics[height=\dimexpr\ht\mytable+\dp\mytable\relax]{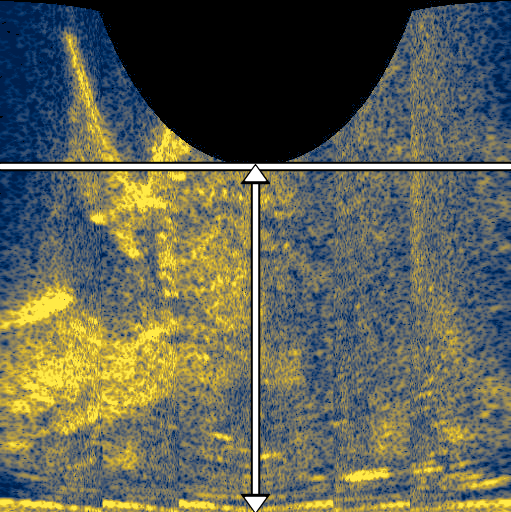}
    \end{minipage}
    \caption{Sonar polar range-line position measured from the bottom of the image to the center-column trough (illustrated right). Each test contains 1,000 randomly sampled frames (4,000 frames total). Std. Dev. (\%) is relative to the median value.}
    \label{tab:num_complete_polar_rows}
\end{figure*}

\Cref{tab:num_complete_polar_rows} summarizes the position of the center-column trough that marks where the polar image ceases to cover the full angular range. Median positions range from 345 to 349 pixels across the four runs, with standard deviations below 1.5 pixels (0.43\% or less). The small within-run variation indicates that this boundary is nearly constant over time.

\section{Human-free usable footage}
The following intervals were selected as footage without people visible in either modality.  Counts below are counts of files/metadata rows actually present between the supplied endpoints; they are not endpoint subtraction alone.

Based on sonar wall-clock timestamps, these intervals span approximately 63.13 minutes (15H), 69.20 minutes (40H), and 47.42 minutes (100H).  The usable sonar subsets contain approximately 85.2\%, 82.5\%, and 79.9\% of each test's extracted sonar frames.  The corresponding RGB subsets contain approximately 88.8\%, 63.8\%, and 82.9\% of the recorded RGB frames.  The lower RGB fraction in 40H is principally a consequence of the deliberately selected human-free endpoint, not evidence of file corruption.

\section{Sonar angular spacing}\label{appendix_angular_spread}
Angular beam spacing is the difference in angle between adjacent sonar beams.
The recorded angle tables were identical
across all sampled frames and all four runs. Individual angular
separations ranged from $0.294114^\circ$ to $0.294122^\circ$, with an average of $0.294117643^\circ$ and 
a maximum deviation from the average spacing of approximately
$4.36 \times 10^{-6}$ degrees. These small differences are consistent
with numerical precision and rounding, indicating effectively
uniform angular spacing.
No frame-to-frame variation in angular sampling was observed in
the inspected frames. This finding describes the recorded angular
coordinates.

\section{Frame rate variation}\label{sec:appendix_frame_rate_decrease}
\begin{figure*}
    \centering
    \begin{subfigure}[t]{0.49\textwidth}
        \centering
        \includegraphics[width=\linewidth]{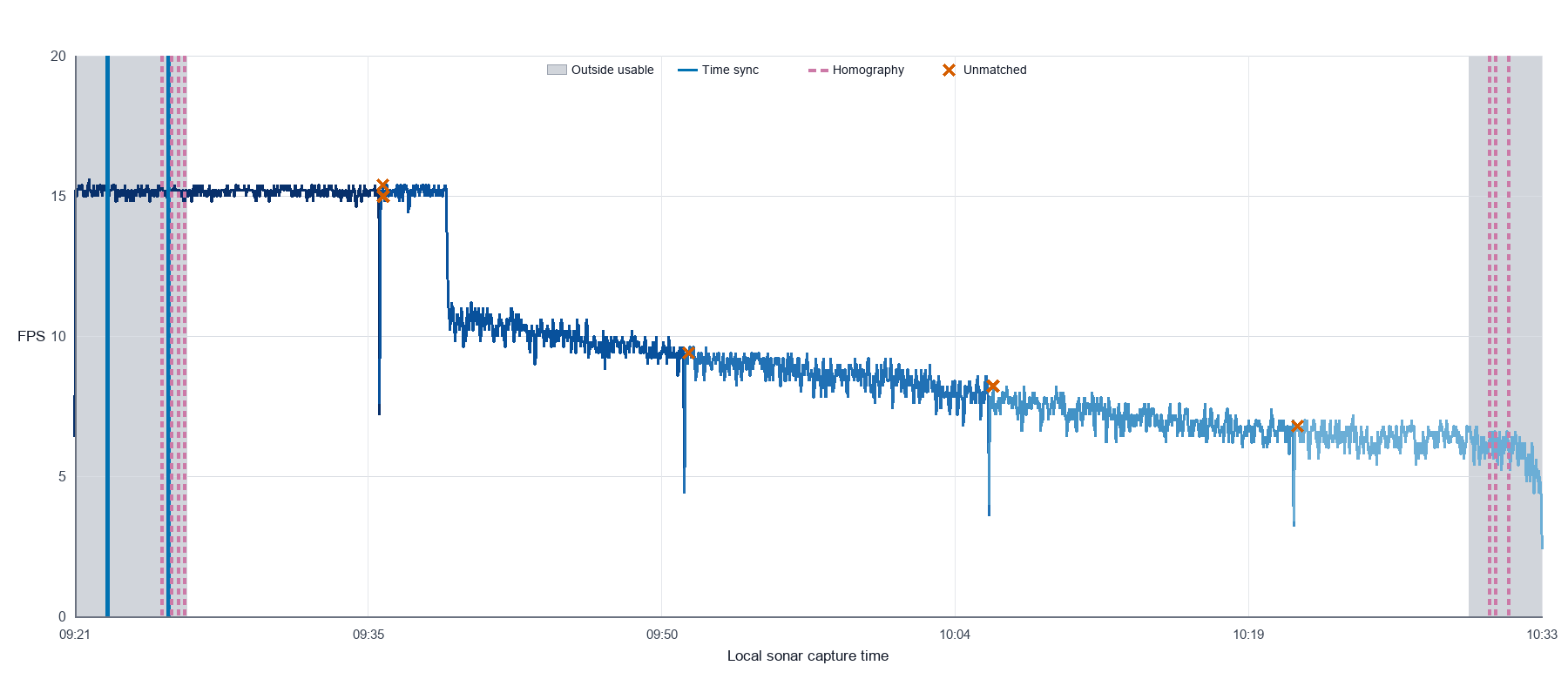}
        \caption{15H usable sonar frames 39,700-111,182}
    \end{subfigure}%
    \hfill 
    \begin{subfigure}[t]{0.49\textwidth}
        \centering
        \includegraphics[width=\linewidth]{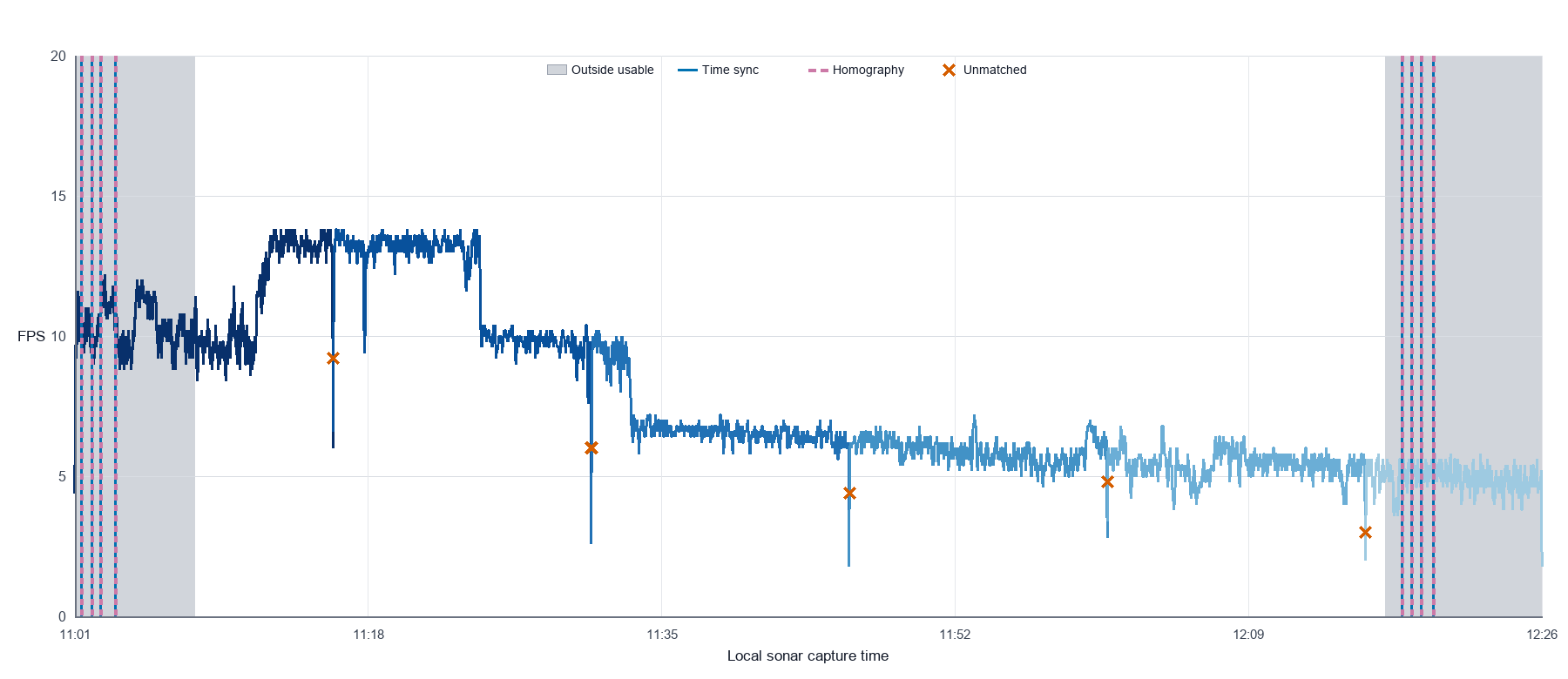}
        \caption{40H usable sonar frames 19,038-75,238}
    \end{subfigure}%
    \\
    \begin{subfigure}[t]{0.49\textwidth}
        \centering
        \includegraphics[width=\linewidth]{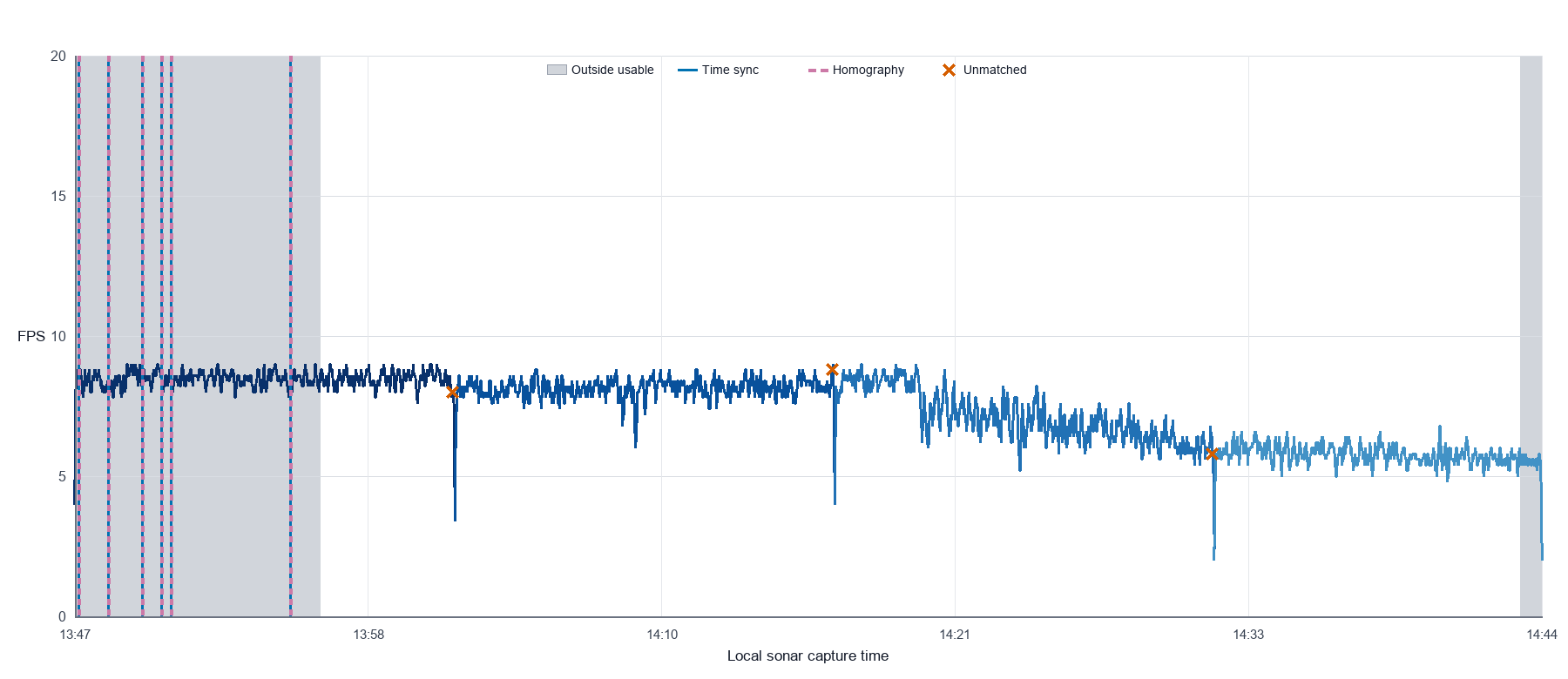}
        \caption{100H usable sonar frames 165,958-191,253}
    \end{subfigure}%
    \hfill 
    \begin{subfigure}[t]{0.49\textwidth}
        \centering
        \includegraphics[width=\linewidth]{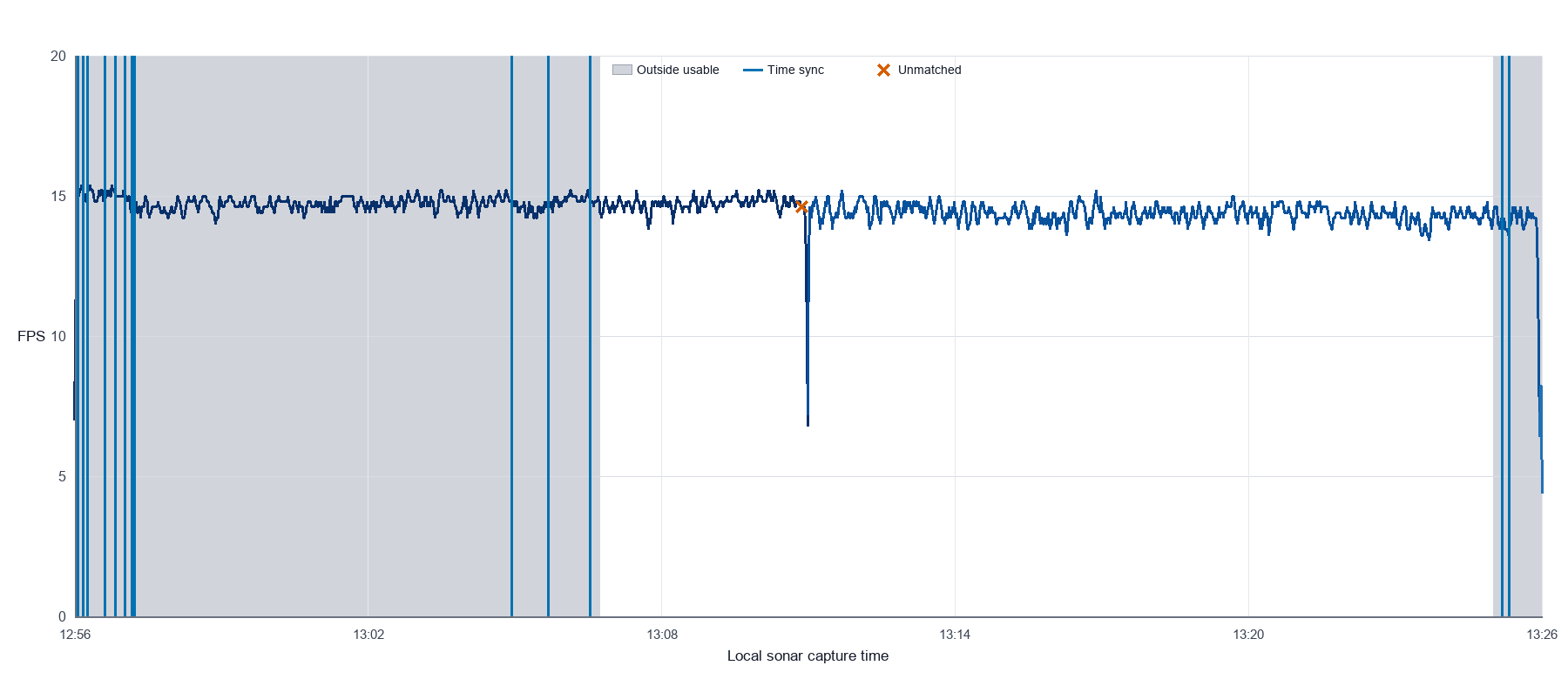}
        \caption{20V usable sonar frames 128,380-148,999}
    \end{subfigure}%
    \caption{Sonar frame rate over time for each recording configuration. Solid blue vertical lines mark time-synchronization events, when a pipe was rapidly inserted into the water to produce a recognizable event in both sonar and RGB video. Dashed pink lines mark frames containing the cross-shaped pipe calibration object used to identify correspondences for homography estimation. These annotations refer to physical objects and actions recorded in the videos, rather than electronic synchronization signals. Gray shading indicates intervals outside the selected usable footage; orange crosses mark sonar frames without an RGB match within 100ms.}
    \label{fig:fps}
\end{figure*}

\begin{figure*}
    \centering
    \includegraphics[width=\linewidth]{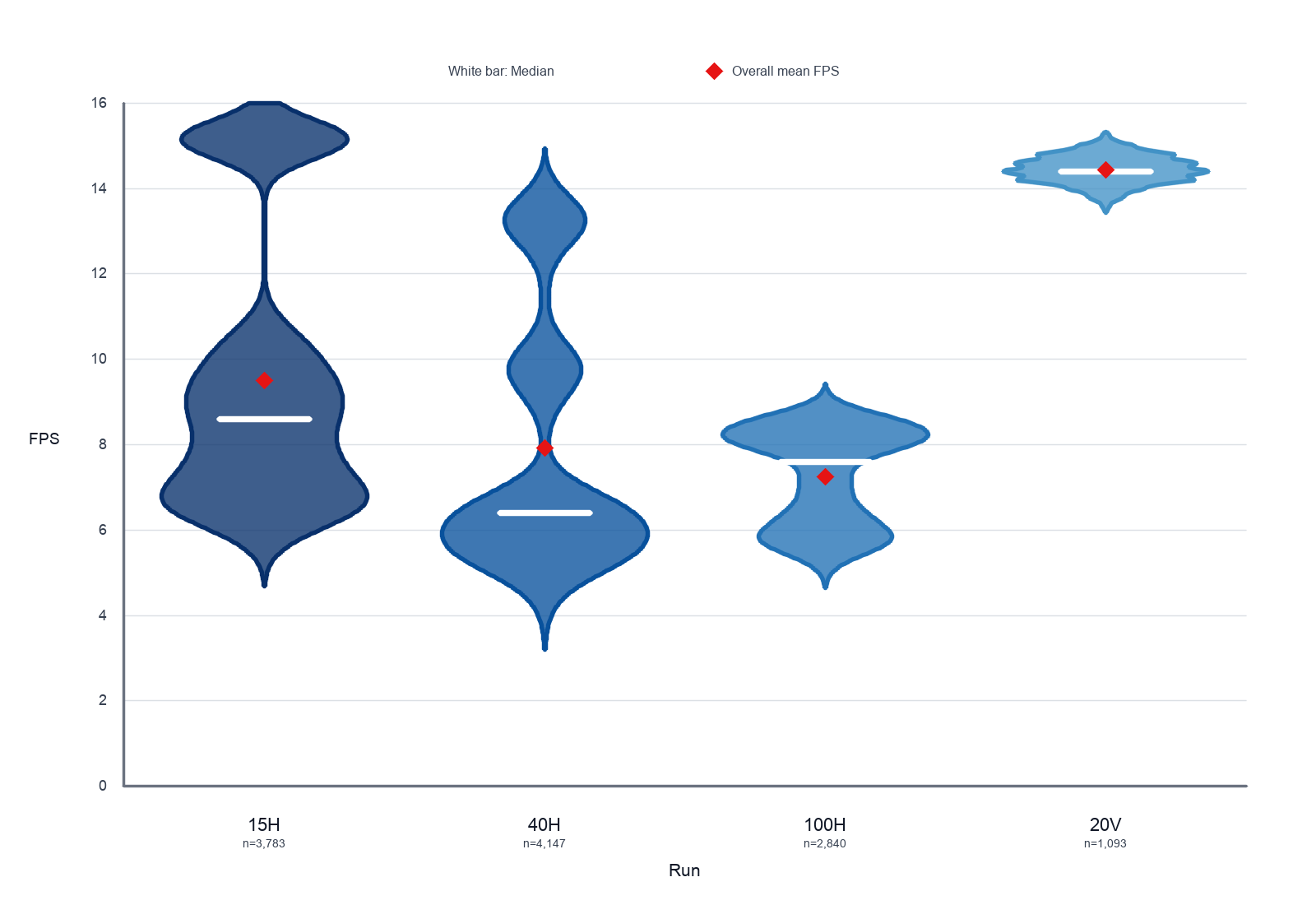}
    \caption{Distributions of sonar frame rates estimated over rolling 5~s windows for the selected usable footage in each recording configuration. Violin width represents the density of frame-rate estimates; white bars mark the medians, and red diamonds mark the overall mean frame rates. The number of estimates is shown below each run label. The horizontal runs exhibit broader, multimodal distributions, whereas 20V has a narrower distribution.}
    \label{fig:violin}
\end{figure*}

The sonar frame rate varies both across configurations and over time, as shown in \cref{fig:fps}. The horizontal runs show sustained decreases during recording, whereas 20V remains comparatively stable. The distributions in \cref{fig:violin} are correspondingly broad and multimodal for the horizontal runs, but concentrated near 14--15 frames per second for 20V. The cause of the decreases has not been determined.

\section{Time synchronization quality}
\label{sec:appendix_time_sync}
RGB wall-clock time was recovered by robust affine fitting of precise camera timestamps to transitions in the integer-second Mac receipt clock.  Capture starts and ends were checked using the number of frames present in their partial wall-clock seconds.  Actual timestamp rows, rather than a fixed 30-fps assumption, are used by the viewers.

\begin{itemize}
  \item \textbf{15H:} one internal RGB timestamp discontinuity occurs at RGB 28,279--28,280 (2.084 s).  The adopted viewer model shifts the pre-discontinuity timeline while preserving the post-discontinuity alignment.  All MOVs still decode exactly the number of frames listed in metadata.
  \item \textbf{40H:} capture 6 contains a documented five-frame drop at RGB 355,191--355,197.  Its timestamp advances by 0.200 s, so alignment after the event is preserved.  No other internal timestamp discontinuity was found.
  \item \textbf{100H:} no internal timestamp or frame-number discontinuity was found.  Capture-boundary gaps are approximately 0.411, 0.392, and 0.407 s and do not accumulate synchronization error because each capture is independently fitted.
  \item \textbf{20V:} no internal timestamp discontinuity was found.  The first RGB recording contains only 22 frames (0.700 s), so it cannot independently observe enough wall-clock-second transitions for an affine fit.  Its start time was recovered by linking its continuous camera PTS clock to the immediately following capture.  Capture-boundary gaps are approximately 1.300, 0.490, and 0.467 s.  Thirteen existing sonar-RGB matched pairs span the two principal captures; their median corrections are $-1.153$ s and $-1.179$ s, with within-capture spreads of 53.2 ms and 18.3 ms.  The dedicated viewer presents raw-polar or warped-XY sonar beside RGB and permits per-capture visual refinement.
\end{itemize}

The timing models locate the nearest recorded RGB timestamp typically within one-half RGB frame (about 16.7 ms), apart from intervals where a modality did not record a new frame.  This timestamp proximity is not, by itself, proof of exposure-level synchronization: the devices lacked a shared hardware trigger, and constant transport latency or rolling-shutter effects may remain.

\section{Shared homography method}\label{sec:appendix_homogrpahy}
For each distance, four point correspondences were selected on each of several synchronized warped-sonar/RGB pairs. The original model used RANSAC with a 12-RGB-pixel threshold and was refined using its consensus points. For this report, each annotated sonar point was mapped exactly from its former raster coordinates into its frame's full-warp coordinates. The original inlier membership was preserved, and the shared matrix was then refitted on those migrated consensus points without new manual annotation. RGB was rotated 180 degrees for 15H, 90 degrees clockwise for 40H, and not rotated for 100H.

The vertical-plane test is intentionally absent from the matrices and error table below.  Its perspective and available correspondences did not support a consistent planar homography, so reporting a pixel or centimeter homography error for it would imply a model that the data does not justify.  Its synchronization remains useful independently of geometric overlay.

For sonar homogeneous point $\mathbf{s}=(x,y,1)^T$, the predicted RGB point is
\begin{equation}
  \widehat{\mathbf{r}} = \pi(H\mathbf{s}), \qquad
  \pi([a,b,c]^T)=[a/c,b/c]^T.
\end{equation}
The reported RGB residual is $\lVert\widehat{\mathbf{r}}-\mathbf{r}\rVert_2$.

\subsection{Homography matrices}
\subsubsection*{15H, RGB rotated 180 degrees}
\scaledmatrix{
H_{15}=\begin{bmatrix}
6.90286426 & 0.339648326 & -1483.03267\\
0.944939818 & 4.64554217 & -540.174085\\
0.00254177561 & -0.000197745457 & 1
\end{bmatrix}
}

\subsubsection*{40H, RGB rotated 90 degrees clockwise}

\scaledmatrix{
H_{40}=\begin{bmatrix}
8.91419579 & -5.86796112 & -446.009924\\
8.59130238 & 5.35463179 & -3894.98968\\
0.00383600753 & -0.00210383850 & 1
\end{bmatrix}
}

\subsubsection*{100H, RGB unrotated}

\scaledmatrix{
H_{100}=\begin{bmatrix}
1.95051635 & -1.73585783 & -2.81100761\\
0.0923878190 & 0.0931856591 & 0.330816802\\
0.000124706315 & -0.00189896824 & 1
\end{bmatrix}
}

\section{Homography accuracy}\label{sec:appendix_homography_accuracy}
\Cref{tab:homography_points} summarizes agreement on the annotated calibration correspondences for the three horizontal runs. Median RGB reprojection errors range from 3.92 to 6.66 pixels, while median sonar-plane residuals increase from 1.70~cm in 15H to 4.93~cm in 100H. These values describe agreement on the calibration points used for fitting, rather than independent localization accuracy.

\subsection{RANSAC-rejected control points}
The stored RANSAC consensus rejected eight of the 84 annotated correspondences: two in 15H, four in 40H, and two in 100H. \Cref{tab:ransac_outliers} identifies these points and reports their residuals after coordinate migration and refitting. Two rejected points have final RGB residuals below the original 12-pixel threshold, illustrating that the preserved consensus labels do not follow directly from the final residuals.

\begin{table*}
\centering
\small
\begin{tabular}{rrrcrr}
\toprule
Run & Sonar frame & RGB frame & Point ID & Sonar residual (px) &  RGB residual (px)\\
\midrule
15H & 38,797 & 32,107 & 3 & 5.17 & 11.36\\
15H & 39,206 & 32,757 & 4 & 6.49 & 16.62\\
40H & 14,609 & 206,883 & 2 & 2.60 & 14.24\\
40H & 76,014 & 342,979 & 3 &  2.34 & 15.13\\
40H & 76,014 & 342,979 & 4 & 2.36 & 15.28\\
40H & 77,530 & 346,337 & 2 & 3.85 & 13.34\\
100H & 160,880 & 501,984 & 1 & 0.95 & 7.99\\
100H & 162,212 & 506,471 & 1 & 1.22 & 12.78\\
\bottomrule
\end{tabular}
\caption{Points rejected by the stored RANSAC consensus. A rejected point may reflect a mis-click, parallax/depth, imperfect timing, or violation of the planar model. Consensus membership was preserved during coordinate migration; it is not equivalent to thresholding the final refined residual alone.}
\label{tab:ransac_outliers}
\end{table*}

\begin{table*}
\centering
\small
\begin{tabular}{rccccccccc}
\toprule

& & & & & \multicolumn{2}{c}{RGB (pixels)} & \multicolumn{3}{c}{Sonar (cm)} \\
\cmidrule(lr){6-7}\cmidrule(lr){8-10}

Run & \# Frames & \# Points & Outliers & Inlier rate & Median & RMSE & Median & RMSE & Inlier RMSE\\
\midrule
15H  & 7 & 28 & 2 & 92.9\% & 3.92 & 6.28 & 1.70 & 2.51 & 2.15\\
40H  & 8 & 32 & 4 & 87.5\% & 5.69 & 7.60 & 3.06 & 4.11 & 3.57\\
100H & 6 & 24 & 2 & 91.7\% & 6.66 & 6.90 & 4.93 & 5.95 & 5.89\\
\bottomrule
\end{tabular}
\caption{Shared-homography agreement. \# Frames and \# Points refer to the number of homography frames and points used.
RGB (pixels) and Sonar (cm) represent the error between the sonar and RGB points projected into the specified image space.
Errors include all points, including outliers, in all columns other than Inlier RMSE.
Outliers are detailed in \cref{tab:ransac_outliers}. The corrected sonar images have a resolution of \(989 \times 512\) pixels for all three runs. The RGB calibration images for 15H and 100H have a resolution of \(1920 \times 1080\) pixels. For 40H, the RGB image is rotated \(90^\circ\), resulting in a calibration-image resolution of \(1080 \times 1920\) pixels.
}
\label{tab:homography_points}
\end{table*}

The centimeter-equivalent residual is computed in the sonar plane, not by assigning a constant physical size to an RGB pixel.  For annotated RGB point $\mathbf{r}$ and its paired sonar point $\mathbf{s}$,
\begin{equation}
 e_{\mathrm{cm}}=100q\left\lVert\pi(H^{-1}\mathbf{r})-\mathbf{s}\right\rVert_2,
\end{equation}
where $q$ is the sonar range-bin scale:
\SI{9.0}{mm/pixel}, \SI{23.6}{mm/pixel}, and \SI{59.6}{mm/pixel}
for the 15-, 40-, and 100H settings.  Because a projective transform has position-dependent scale, this inverse mapping is preferable to multiplying the RGB residual by one global RGB-pixel scale.
The original RANSAC fit used a **12-RGB-pixel reprojection threshold**.

For each candidate homography, a sonar point was projected into the RGB image. Its reprojection error was calculated as:

\[
e_{\mathrm{RGB}}
=
\left\|
\pi(H\mathbf{s})-\mathbf{r}
\right\|_2.
\]

A correspondence was considered an inlier when this error was within approximately 12 RGB pixels under the RANSAC candidate model; otherwise, it was rejected from that model’s consensus set.

Importantly, the table retains the original RANSAC classifications after migrating the sonar points into the corrected warp and refitting the homography. Therefore, the final displayed residual alone does not determine whether a point is listed as rejected. A preserved outlier can now have a final residual below 12 pixels.

\end{document}